%% file: iclr2027_conference.tex
\documentclass{article} 
\usepackage{iclr2027_conference,times}

\input{math_commands.tex}

\usepackage{hyperref}
\usepackage{url}
\usepackage{amsmath}
\usepackage{amssymb}
\usepackage{amsthm}
\usepackage{booktabs}
\usepackage{graphicx}
\usepackage{placeins}

\newcommand{\Adv}{\mathcal{A}}
\newtheorem{proposition}{Proposition}

\title{GraftOOD: Grafting Out-of-Distribution Exemplars into Advantage-Weighted Rectification for Video Reward Optimization}

\author{%
\textbf{Zheming Fu$^{1,2,3}$}\thanks{These authors contributed equally.}, \hspace{.3em}
\textbf{Ruizhe He$^{3,4}$}\footnotemark[1], \hspace{.3em}
\textbf{Wenbin Ning$^{1}$}\footnotemark[1], \hspace{.3em}
\textbf{Xiaoxiao Ma$^{3,5}$}, \hspace{.3em}
\textbf{Xuan Yang$^{3}$},\\
\textbf{Lianrui Mu$^{3,4}$}, \hspace{.3em}
\textbf{Haoyang Huang$^{3}$}, \hspace{.3em}
\textbf{Chang Liu$^{2}$}\thanks{Corresponding authors.}, \hspace{.3em}
\textbf{Lei Wang$^{1}$}\footnotemark[2], \hspace{.3em}
\textbf{Siming Fu$^{3}$}\footnotemark[2]\\[1ex]
$^{1}$ Beihang University, $^{2}$ Zhongguancun Academy, $^{3}$ Joy Future Academy,\\
$^{4}$ Zhejiang University, $^{5}$ University of Science and Technology of China
}

\iclrfinalcopy

\begin{document}

\maketitle
\lhead{}

\begin{abstract}
Video reward optimization can increase proxy rewards while reducing video dynamics,
a mismatch we identify as \emph{motion collapse}, including under a video-level
motion-quality reward. We propose \textbf{GraftOOD}, which replaces part of the
online rollouts in selected prompt groups with out-of-distribution video exemplars.
A shared velocity-regression objective combines group-relative reward rectification
with sequence-level exemplar supervision, requiring neither exemplar generation
trajectories nor an additional supervised-loss coefficient. We characterize its
pointwise optimal field as a source-posterior-weighted combination of the two
source-specific target fields. We construct \textbf{GraftSet-Video}, containing
9,588 format-matched exemplars for 2,397 prompts, to supply external supervision.
On Wan2.1-1.3B, GraftOOD raises Dynamic Degree from 4.86 to 44.41 relative to
reward-only FlowAWR, while HPSv3 decreases from 8.10 to 7.32 and VisionReward
remains close. At the main setting, GraftOOD leads nine of ten metrics among
three exemplar-augmented methods. Ablations of exemplar frequency and within-group
count reveal dynamics--reward trade-offs rather than uniform gains from increasing
exemplar usage. These results support external exemplar supervision as a complement
to scalar reward feedback, without equating higher measured dynamics with coherent,
visually plausible motion in the generated videos.
Code is available at \href{https://github.com/Fuzm7/FlowAWR}{GraftOOD}.
\end{abstract}

\section{Introduction}

Reinforcement learning has become a central component of generative-model
post-training, enabling learning from human preferences
\citep{christiano2017deep,ouyang2022training} and verifiable outcomes
\citep{shao2024deepseekmath}. Unlike supervised imitation alone, it lets a
model sample its own outputs, receive feedback, and concentrate updates on
behaviors that satisfy the specified criteria. This principle extends to
visual generation, where learned evaluators guide improvements in aesthetic
quality and prompt correspondence
\citep{black2024ddpo,fan2023dpok,xue2025dancegrpo,zheng2025diffusionnft}.
These advances depend not only on the optimization algorithm, but also on
how well the reward represents the preferences that define the desired outputs.

\begin{figure}[!t]
    \centering
    \includegraphics[width=\linewidth]{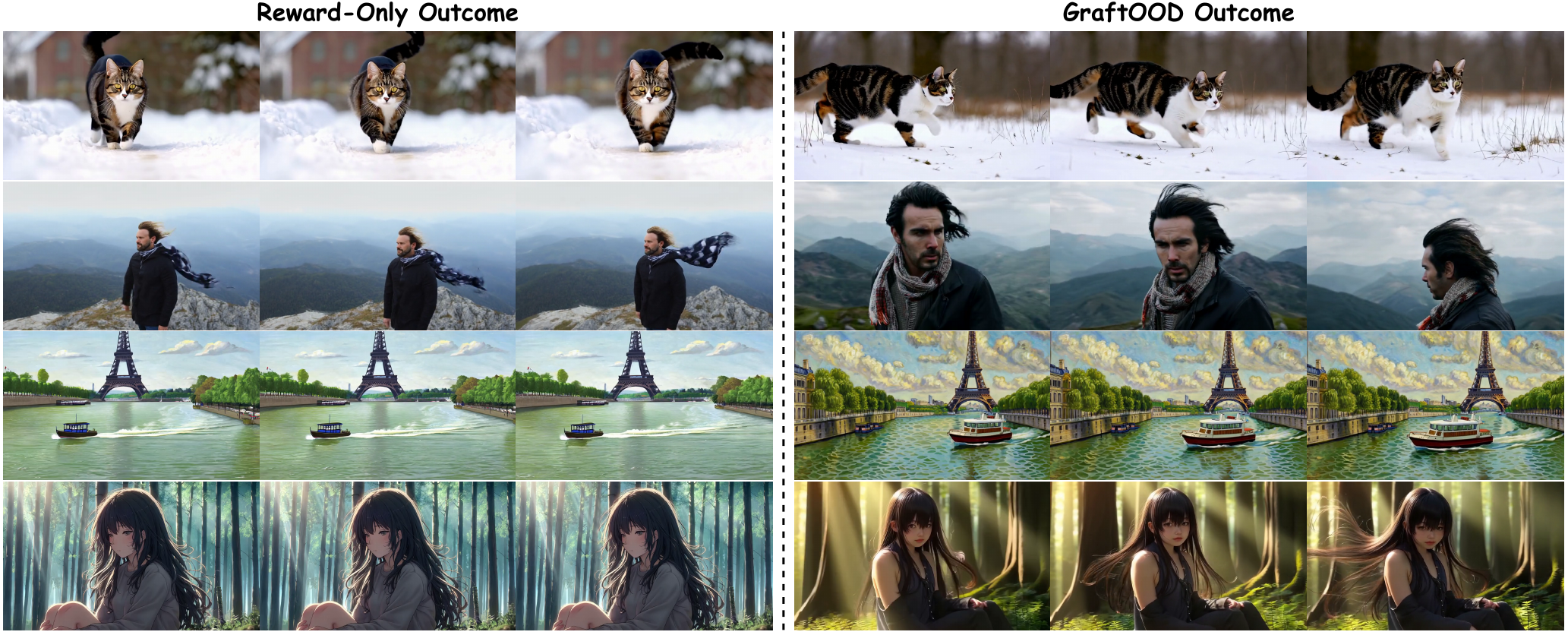}
    \vspace{-4pt}
    \caption{\textbf{Reward-only optimization versus GraftOOD.}
    Same-prompt sequences from reward-only optimization (left) and GraftOOD
    (right), with frames in temporal order.}
    \label{fig:graftood_rl_compare}
    \vspace{-0.08in}
\end{figure}

For video generation, a proxy reward must account for both individual-frame
quality and how content evolves over time. Optimizing this proxy favors
high-scoring outputs, which need not satisfy all of these preferences. In our
experiments, optimizing frame-aggregated HPSv3 \citep{ma2025hpsv3,team2025longcat}
raises its score while reducing Dynamic Degree \citep{huang2024vbench}.
Optimizing VideoAlign's Motion Quality \citep{liu2026improving} likewise
raises its reward while reducing Dynamic Degree. We refer to this mismatch
between rising rewards and declining video dynamics as \emph{motion collapse}.
These observations are consistent with reward exploitation: a policy may
increasingly favor high-reward outputs whose temporal behavior is undesirable.

Existing methods improve reward optimization through policy gradients or
regression objectives \citep{liu2026flow,li2025mixgrpo,he2026flash}, but
improving a proxy objective does not by itself resolve its mismatch with the
desired outputs. Combining rewards is one response, yet it requires choosing
which attributes to score and how to balance them. In our experiments, adding
a RAFT-based motion reward \citep{teed2020raft} to HPSv3 mitigates but does not
prevent the decline in Dynamic Degree. Rather than relying solely on further
reward design, external videos can provide examples of outputs we want the
policy to learn from. This raises the question:
\textbf{Can an external exemplar distribution guide online reward optimization
toward desired video attributes that the reward does not adequately constrain?}

DiffusionNFT \citep{zheng2025diffusionnft} optimizes rewards through velocity
regression rather than likelihood ratios over generation trajectories.
FlowAWR \citep{fu2026flowawr} further derives advantage-weighted velocity
rectification. This regression view provides a shared interface for online
reward optimization and external supervision: both update the same velocity
network, but they differ in the sample sources and regression targets that
define each update.

We introduce \textbf{GraftOOD}, which replaces part of the online rollouts
in selected prompt groups with out-of-distribution (OOD) video exemplars.
Online targets retain group-relative reward rectification, while exemplars
provide sequence-level supervision without their generating trajectories.
Both enter one regression objective without an additional supervised-loss
coefficient. We characterize its pointwise optimal field as a
source-posterior-weighted combination of the online target field and the
exemplar reference field (Figure~\ref{fig:graftood_overview}, top).
This analysis describes the regression target, rather than guaranteeing
properties of generated videos. Figure~\ref{fig:graftood_rl_compare} gives
same-prompt qualitative comparisons of video sequences generated with GraftOOD
and reward-only optimization over multiple frames.

We construct \textbf{GraftSet-Video} from public video reward-model resources,
supplementing selected prompt--video pairs with outputs from external
generators, including Seedance and Kling. The corpus contains 9,588 videos
for 2,397 prompts outside the original policy-training set, with four
exemplars per prompt and the policy's resolution and frame count.

On Wan2.1-1.3B, GraftOOD uses an exemplar-prompt ratio of 0.167 and four
exemplars per group in the main comparison. Relative to FlowAWR, it raises
Dynamic Degree from 4.86 to 44.41, while HPSv3 decreases from 8.10 to 7.32
and VisionReward remains close. GraftOOD leads nine of ten
metrics in this comparison with FlashOOD and NFTOOD. Ablations of both
exemplar frequency and within-group count reveal dynamics--reward trade-offs,
not uniform improvements from increasing exemplar usage. Higher Dynamic
Degree alone does not establish better motion quality.

Our contributions are as follows:

\begin{itemize}
    \item We identify \emph{motion collapse} in video reward optimization:
    increasing proxy rewards can coincide with declining video dynamics,
    even with a video-level motion-quality reward. These observations
    motivate sequence-level supervision beyond scalar reward feedback.

    \item We propose GraftOOD, which grafts OOD video exemplars into
    group-relative online optimization through a shared velocity-regression
    objective, without an additional supervised-loss coefficient. We
    characterize its pointwise optimal field as a source-posterior-weighted
    combination of the two source-specific target fields.

    \item We construct GraftSet-Video, comprising 9,588 exemplars for 2,397
    prompts, and evaluate exemplar-augmented optimization on Wan2.1-1.3B.
    GraftOOD improves dynamics over reward-only FlowAWR and leads
    nine of ten metrics among the three OOD methods at the main setting,
    while ratio and count ablations reveal dynamics--reward trade-offs.
\end{itemize}

\section{Background}

\subsection{Trajectory-Based Policy Optimization}
\label{subsec:bg_grpo}

Flow models generate samples by integrating the deterministic ODE
$\mathrm{d}x_t=v_\theta(x_t,c,t)\,\mathrm{d}t$. For policy-gradient training,
FlowGRPO \citep{liu2026flow} instead adopts a marginal-preserving SDE,
whose discretization provides tractable transition likelihoods.

For each prompt $c$, the terminal rewards of $G$ on-policy trajectories
are normalized using the prompt-group mean $\overline{R}_c$ and the reward
standard deviation over the current rollout batch to form advantages:
\begin{equation}
    \Adv^{(i)}
    =\frac{R(c,x_1^{(i)})-\overline{R}_c}
    {\operatorname{Std}(R)+\epsilon}.
    \label{eq:grpo_advantage}
\end{equation}
These advantages weight the trajectory log-probabilities in the on-policy
surrogate objective:
\begin{equation}
    \mathcal{L}_{\mathrm{GRPO}}(\theta)
    =-\mathbb{E}_{c,\{\tau^{(i)}\}\sim p_\theta}
    \left[
    \frac{1}{G}\sum_{i=1}^{G}
    \Adv^{(i)}
    \sum_{k=0}^{K-1}
    \log p_\theta(x_{t_{k+1}}^{(i)}\mid x_{t_k}^{(i)},c)
    \right].
    \label{eq:onpolicy_grpo}
\end{equation}
where the likelihood terms are evaluated along the sampled SDE paths.

\subsection{Reward-Guided Velocity Regression}
\label{subsec:bg_regression}

Regression-based methods instead incorporate rewards into velocity regression
without evaluating trajectory likelihoods. Under Rectified Flow
\citep{lipman2022flow,liu2023flow}, a terminal video latent $x_1$ and fresh
noise $x_0\sim p_0=\mathcal{N}(0,I)$ define the training state and conditional
velocity target:
\begin{equation}
    x_t=(1-t)x_0+t x_1,
    \qquad
    u_t(x_t\mid x_1)=x_1-x_0
    =\frac{x_1-x_t}{1-t}.
    \label{eq:rectified_flow}
\end{equation}

DiffusionNFT \citep{zheng2025diffusionnft} uses the normalized terminal reward
$r\in[0,1]$ to weight regression toward positive and negative corrections of
a fixed reference field $v^{\mathrm{old}}$. For fixed $\beta>0$, its objective
can be rewritten, omitting spatiotemporal arguments, as
\begin{equation}
\begin{aligned}
    \mathcal{L}_{\mathrm{NFT}}(\theta)=\beta^2\mathbb{E}\Big[&
    r \left\|v_\theta-\left(v^{\mathrm{old}}
    +\frac{u_t-v^{\mathrm{old}}}{\beta}\right)\right\|_2^2 +(1-r)\left\|v_\theta-\left(v^{\mathrm{old}}
    -\frac{u_t-v^{\mathrm{old}}}{\beta}\right)\right\|_2^2\Big].
\end{aligned}
\label{eq:nft_loss}
\end{equation}

Starting from the optimal terminal distribution under KL-regularized reward
maximization, FlowAWR \citep{fu2026flowawr} derives its intermediate marginals
and corresponding velocity field:
\begin{equation}
    v^\star(x_t,c,t)=v^{\mathrm{old}}(x_t,c,t)
    +\mathbb{E}_{x_1\sim p^{\mathrm{old}}(\cdot\mid x_t,c)}
    \left[\frac{\exp(R(c,x_1)/\gamma)}{\phi(x_t,c,t)}
    (u_t-v^{\mathrm{old}})\right].
    \label{eq:awr_background_field}
\end{equation}
Since $\mathbb{E}_{p^{\mathrm{old}}(\cdot\mid x_t,c)}
[u_t-v^{\mathrm{old}}]=0$, subtracting the normalized weight's mean yields the advantage
\begin{equation}
    \Adv_t(x_1,x_t,c)=\frac{\exp(R(c,x_1)/\gamma)}{\phi(x_t,c,t)}-1,
    \label{eq:awr_population_advantage}
\end{equation}
with $\gamma>0$ and
$\phi(x_t,c,t)=\mathbb{E}_{x_1'\sim p^{\mathrm{old}}(\cdot\mid x_t,c)}
[\exp(R(c,x_1')/\gamma)]$.
Following the reduction in conditional flow matching
\citep{lipman2022flow}, regression toward $v^\star$ uses sample targets:
\begin{equation}
    \mathcal{L}_{\mathrm{AWR}}(\theta)
    =\mathbb{E}_{\substack{x_1\sim p^{\mathrm{old}}(\cdot\mid c)\\
    t,\,x_0\sim p_0}}
    \left[\left\|v_\theta-
    \left(v^{\mathrm{old}}+\Adv_t(x_1,x_t,c)
    \left(u_t-v^{\mathrm{old}}\right)\right)
    \right\|_2^2\right],
    \label{eq:awr_loss}
\end{equation}
where GRPO-style estimation gives
$\Adv_t\approx\widehat{\Adv}_0^{(i)}
=\exp(R_i/\gamma)/[G^{-1}\sum_{j=1}^{G}\exp(R_j/\gamma)]-1$.

\begin{figure}[!t]
    \centering
    \includegraphics[width=\textwidth]{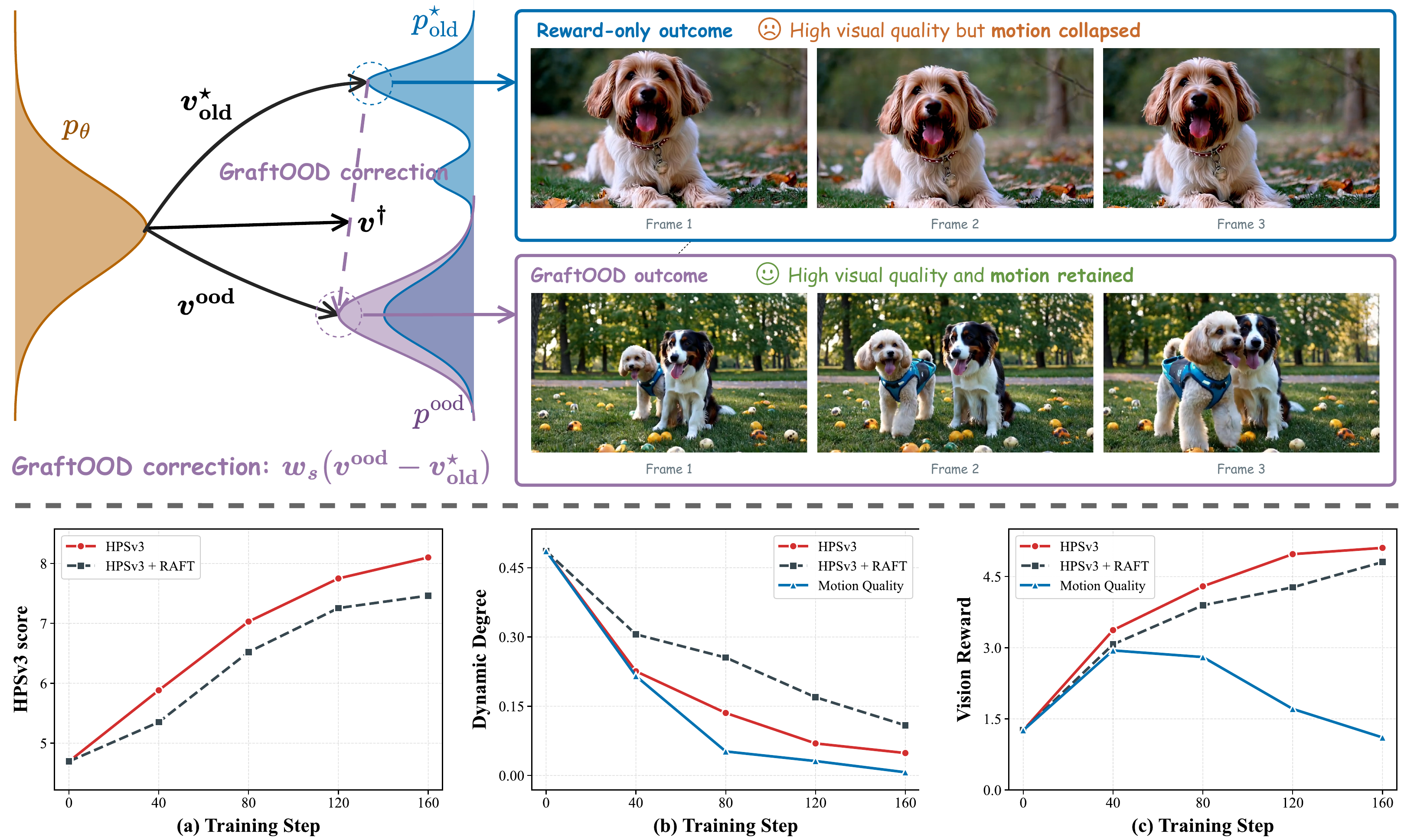}
    \vspace{-0.2in}
    \caption{\textbf{GraftOOD motivation and overview.}
    \textbf{Top:} exemplar-guided correction
    $w_s(v^{\mathrm{ood}}-v_{\mathrm{old}}^\star)$ and motion examples.
    \textbf{Bottom:} \textbf{(a)} HPSv3, \textbf{(b)} Dynamic Degree, and
    \textbf{(c)} VisionReward during reward optimization. Adding RAFT to HPSv3
    mitigates but does not prevent declining dynamics.}
    \label{fig:graftood_overview}
    \vspace{-0.2in}
\end{figure}

\section{Video Reinforcement with OOD Distribution Grafting}
\label{sec:method}

\subsection{Problem Setup}
\label{subsec:dual_source}

Consider a video policy conditioned on a prompt $c$ and updated using a
terminal reward $R(c,x_1)$. Online samples provide training experience, while
the reward determines which outputs to encourage or suppress. As a proxy for
the desired video attributes, the reward may also favor undesirable outputs.
Optimizing it can then concentrate the policy on a high-reward region that
does not reflect our preferences for visual appearance and temporal behavior.

Figure~\ref{fig:graftood_overview}(a,b) illustrates this mismatch: optimizing
HPSv3 raises its score while Dynamic Degree declines. Optimizing VideoAlign
Motion Quality likewise raises its reward while reducing Dynamic Degree
(Appendix~\ref{app:additional_results}). Adding a RAFT-based motion reward
mitigates but does not prevent the decline observed with HPSv3 in the same
panel. Meanwhile, Figure~\ref{fig:graftood_overview}(c) shows that HPSv3 training
increases VisionReward \citep{xu2026visionreward}, an evaluator not used for
training. These results motivate \textbf{supplementing reward optimization
rather than replacing it}: improvement on another evaluator coexists with
loss of dynamics in the videos produced during reward optimization.

Additional reward terms revise how online samples are scored, requiring choices
about the attributes to measure and their relative weights. We instead introduce
external videos as a distributional reference alongside reward feedback. They
supply sequences for the policy to learn from without first generating those
examples itself. The selected exemplars provide direct sequence-level targets
to guide policy learning toward desired video attributes.

We construct the OOD exemplar distribution $p^{\mathrm{ood}}(x_1\mid c)$
from prompt--video pairs selected from public video reward-model resources
\citep{liu2026improving,xu2026visionreward,he2024videoscore,he2025videoscore2,tong2025mjvideo},
supplemented with outputs from external generators such as Seedance and Kling.
These videos are selected to provide higher-quality references for the base
policy. Their paired prompts lie outside the original policy-training set,
and the videos match the policy's training format. OOD is defined relative
to this policy setting, not necessarily to the reward model's training data.
The policy also generates fresh rollouts on these prompts for reward-relative
updates. The exemplars provide clean terminal latents, but not their generating
trajectories. \emph{How can we incorporate these external video references into
online reward optimization, so that policy updates are guided by both reward
feedback and the sequence-level attributes of the selected exemplars?}

\subsection{Group-Wise Exemplar Grafting}
\label{subsec:group_grafting}

\paragraph{Trajectory-based optimization with exemplar supervision.}
These terminal-only exemplars cannot be inserted directly as on-policy
samples in \Eqref{eq:onpolicy_grpo}. We therefore supplement online policy
optimization with supervised fine-tuning (SFT), which constructs regression
inputs by adding noise to the encoded exemplar latents. We use Flash-GRPO's
single-timestep scheme \citep{he2026flash} to train the online and exemplar
objectives at a matched noise level. For each prompt group $s$, we randomly
select a timestep $t_s$, compute the online policy loss from the corresponding
stochastic transition, and construct the exemplar SFT loss at the same noise
level. Specifically, each encoded exemplar $x_1^{(j)}$ is independently noised
as $x_{t_s}^{(j)}=(1-t_s)x_0^{(j)}+t_s x_1^{(j)}$, with $x_0^{(j)}\sim p_0$.
The network takes $x_{t_s}^{(j)}$ as input and regresses
$u_{t_s}^{(j)}=u_{t_s}(x_{t_s}^{(j)}\mid x_1^{(j)})$, defined in
\Eqref{eq:rectified_flow}. Using the on-policy form of
\Eqref{eq:onpolicy_grpo}, we combine the single-transition policy objective
with exemplar SFT at the shared timestep $t_s$:
\begin{equation}
    \widehat{\mathcal{L}}_s
    =-\frac{1}{m_s}\sum_{i=1}^{m_s}
    \Adv^{(i)}\log p_\theta\!\left(
    x_{t_s+1}^{(i)}\mid x_{t_s}^{(i)},c_s\right)
    +\frac{\lambda_{\mathrm{ood}}}{n_{\mathrm{ood}}}
    \sum_{j=1}^{n_{\mathrm{ood}}}
    \left\|v_\theta(x_{t_s}^{(j)},c_s,t_s)-u_{t_s}^{(j)}\right\|_2^2.
\label{eq:flash_ood_loss}
\end{equation}
For the $m_s$ online rollouts, $x_{t_s+1}^{(i)}$ denotes the latent
obtained from $x_{t_s}^{(i)}$ by one discretized SDE denoising step.
The advantages $\Adv^{(i)}$ are computed exclusively from online-rollout
rewards using the normalization in \Eqref{eq:grpo_advantage}.
The coefficient $\lambda_{\mathrm{ood}}$ controls the relative weight
of exemplar SFT. We refer to this exemplar-augmented formulation as
\textbf{FlashOOD}.

\paragraph{Reinforcement learning through terminal fine-tuning.}
Off-policy RL through terminal fine-tuning provides a direct regression
interface for these exemplars. In the schemes considered here
\citep{zheng2025diffusionnft,fu2026flowawr}, sampling uses an old policy whose
parameters are an exponential moving average (EMA) of the trainable parameters.
Training re-noises the collected clean latents for reward-guided velocity
regression. For an external supervision term, terminal sampling in the outer
expectation changes from $x_1\sim p^{\mathrm{old}}(\cdot\mid c)$ to
$p^{\mathrm{ood}}(\cdot\mid c)$, while the online term is retained.
When using normalized rewards, we assign the OOD references the theoretically maximal reward.

For example, DiffusionNFT uses the reward-weighted positive and negative
corrections in \Eqref{eq:nft_loss}. Let
$\widehat{\mathcal{L}}_{\mathrm{NFT},s}$ denote its empirical average over the
$m_s$ old-policy rollouts. For exemplars, $r=1$ removes the negative term.
With $v^{\mathrm{old},(j)}=v^{\mathrm{old}}(x_{t_s}^{(j)},c_s,t_s)$,
we combine the online and exemplar mean losses with equal weights
at the shared $t_s$:
\begin{equation}
    \widehat{\mathcal{L}}_{\mathrm{NFT},s}^{\mathrm{ood}}
    =\widehat{\mathcal{L}}_{\mathrm{NFT},s}
    +\frac{1}{n_{\mathrm{ood}}}
    \sum_{j=1}^{n_{\mathrm{ood}}}
    \left\|(1-\beta)v^{\mathrm{old},(j)}
    +\beta v_\theta(x_{t_s}^{(j)},c_s,t_s)-u_{t_s}^{(j)}\right\|_2^2.
\label{eq:nft_ood_loss}
\end{equation}
The online objective weights both corrections by $r_i$ and $1-r_i$,
whereas exemplars receive only the positive correction. And under the default setting $\beta=1$ \citep{zheng2025diffusionnft}, the OOD
term reduces to standard SFT, regressing $v_\theta(x_{t_s}^{(j)},c_s,t_s)$
directly toward $u_{t_s}^{(j)}$. We refer to this DiffusionNFT formulation
with positive-only exemplar supervision as \textbf{NFTOOD}.

\begin{figure}[!t]
    \centering
    \includegraphics[width=\textwidth]{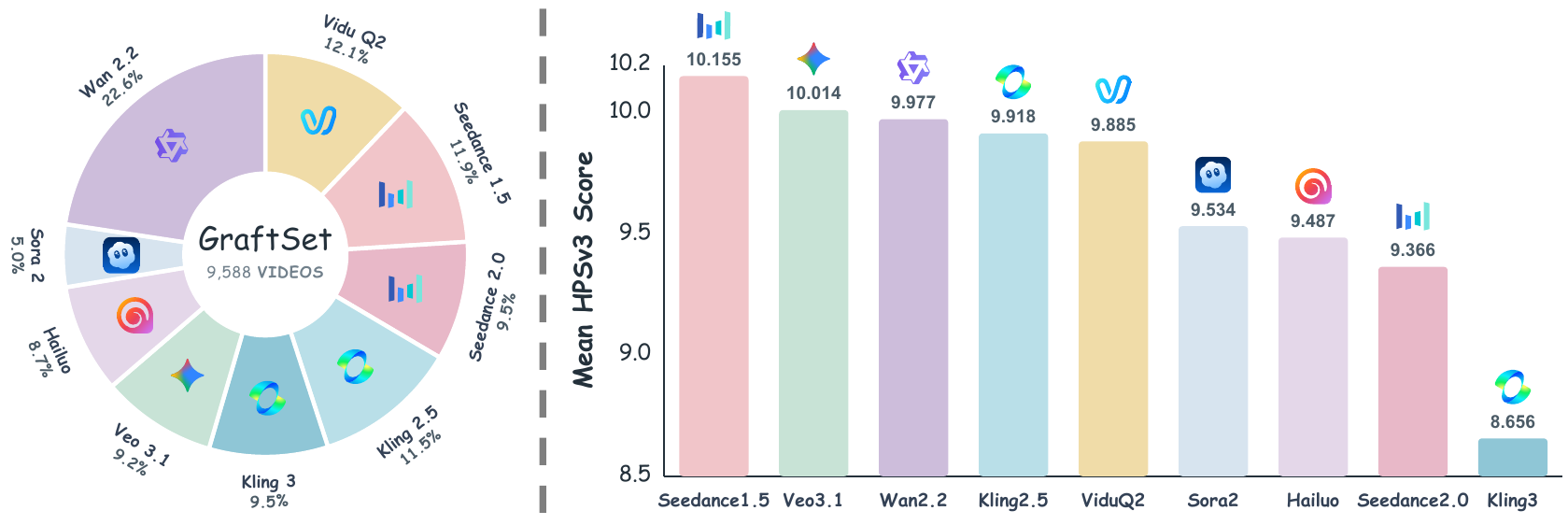}
    \vspace{-4pt}
    \caption{\textbf{GraftSet-Video composition and scores.}
    Source proportions among 9,588 exemplars (left) and mean HPSv3 by source,
    ranked in descending order (right). Colors match across panels.}
    \label{fig:graftset_sources}
    \vspace{-0.08in}
\end{figure}

\paragraph{Group-relative velocity rectification and exemplar grafting.}
Each online sample provides its own residual direction
$u_t^{(i)}-v^{\mathrm{old},(i)}$, whose contribution to the regression target
is controlled by its group-relative coefficient $\widehat{\Adv}_0^{(i)}$.
Assume an OOD reference policy consistent with $p^{\mathrm{ood}}$
and the same forward noising path, with velocity field
$v^{\mathrm{ood}}(x_t,c,t)$. Applying \Eqref{eq:awr_background_field}
to this reference without centering the normalized reward coefficient gives
\begin{equation}
    v_{\mathrm{ood}}^\star=v^{\mathrm{ood}}
    +\mathbb{E}_{p^{\mathrm{ood}}(\cdot\mid x_t,c)}
    \left[\Adv_t^{\mathrm{ood}}(u_t-v^{\mathrm{ood}})\right].
    \label{eq:ood_rectified_field}
\end{equation}
We regard the selected OOD distribution as a desired reference, accepting
its exemplars without further training-time reward differentiation.
Assigning a common reference reward $R^\star$ gives
\begin{equation}
    \Adv_t^{\mathrm{ood}}(x_1,x_t,c)
    :=\frac{\exp(R^\star/\gamma)}
    {\mathbb{E}_{x_1'\sim p^{\mathrm{ood}}(\cdot\mid x_t,c)}
    [\exp(R^\star/\gamma)]}=1.
    \label{eq:ood_unit_coefficient}
\end{equation}
For an exemplar prompt $c_s$, let $\delta=n_{\mathrm{ood}}/G$ be the exemplar
fraction in a group of size $G$. Following the conditional flow matching
reduction used in \Eqref{eq:awr_loss}, we obtain our \textbf{GraftOOD} objective
by combining sample-level regression under the two terminal distributions:
\begin{equation}
\begin{aligned}
    \mathcal{L}_{\mathrm{GraftOOD}}(\theta\mid c_s)
    ={}&(1-\delta)\mathbb{E}_{\mathrm{old}}^{c_s}
    \left[\left\|v_\theta-
    \left[v^{\mathrm{old}}+\widehat{\Adv}_0
    (u_t-v^{\mathrm{old}})\right]\right\|_2^2\right]\\
    &+\delta\mathbb{E}_{\mathrm{ood}}^{c_s}
    \left[\left\|v_\theta-
    \left[v^{\mathrm{ood}}+\Adv_t^{\mathrm{ood}}
    (u_t-v^{\mathrm{ood}})\right]\right\|_2^2\right].
\end{aligned}
\label{eq:graftood_source_objective}
\end{equation}
where the two expectations sample terminals from $p^{\mathrm{old}}(\cdot\mid c_s)$
and $p^{\mathrm{ood}}(\cdot\mid c_s)$ under the same
re-noising timestep. In the OOD term, $\Adv_t^{\mathrm{ood}}=1$
cancels the reference field, making the exemplar loss directly computable
from re-noised exemplars.
And the online term retains group-relative rectification through
$\widehat{\Adv}_0$, which is estimated from online rollouts only. Then \textbf{what velocity field do online reward rectification and exemplar supervision jointly define on an exemplar prompt?}

\begin{proposition}[Exemplar-guided velocity rectification]
\label{prop:convex}
The pointwise optimal velocity field of
$\mathcal{L}_{\mathrm{GraftOOD}}(\theta\mid c_s)$ satisfies
\begin{equation}
    v^\dagger=(1-w_s)v_{\mathrm{old}}^\star+w_s v^{\mathrm{ood}},
    \label{eq:pointwise_optimum}
\end{equation}
where $v_{\mathrm{old}}^\star$ is the reward-rectified old-policy target mean
conditional on $(x_t,c_s,t)$, including group randomness. The proof is given
in Appendix~\ref{app:proof_convex}. The exemplar-source posterior is
\begin{equation}
    w_s(x_t,t)=\frac{\delta p_t^{\mathrm{ood}}(x_t\mid c_s)}
    {(1-\delta)p_t^{\mathrm{old}}(x_t\mid c_s)
    +\delta p_t^{\mathrm{ood}}(x_t\mid c_s)}.
    \label{eq:prompt_source_posterior}
\end{equation}
\end{proposition}
Relative to the online target under the same group-sampling scheme,
exemplars contribute $w_s(v^{\mathrm{ood}}-v_{\mathrm{old}}^\star)$.
The target therefore combines the reward-rectified old-policy field with
an external reference rather than relying on reward rectification alone.
Their relative weight depends on both the exemplar fraction and the
noised-state densities.

\subsection{Practical Implementation}
\label{subsec:ood_corpus}

\paragraph{GraftSet-Video.}
We construct \textbf{GraftSet-Video} with exactly four video exemplars per
prompt, yielding 9,588 videos for 2,397 prompts. We first select videos from
public video reward-model training and evaluation resources, including those
accompanying
\citet{liu2026improving,xu2026visionreward,he2024videoscore,he2025videoscore2,tong2025mjvideo},
using HPSv3 scores \citep{ma2025hpsv3}. Keeping the selected videos, we
supplement prompts with fewer than four exemplars using additional outputs
from external generators, including Seedance
\citep{seedance2025seedance,seedance2026seedance} and Kling
\citep{team2025kling}, to meet this four-exemplar requirement. These generators are chosen to provide
higher-quality video references for the base policy beyond its own online rollouts.

HPSv3-based selection is performed offline, inherits HPSv3's limited motion
sensitivity, and does not enter training-time advantage estimation. Videos
contain 81 frames at $832\times480$ and 16 fps, encoded with the policy's VAE
without further resampling. The prompts are disjoint from the original
19,700 training prompts and 300 evaluation prompts.
Source resources may overlap reward-model training data; OOD refers to the
policy setting above. Figure~\ref{fig:graftset_sources} shows the source
composition and the mean HPSv3 score for each source;
Appendix~\ref{app:corpus_details} provides dataset details.

\paragraph{Exemplar scheduling and allocation.}
We control exemplar usage through the OOD prompt ratio $\rho$ and the
within-group exemplar count $n_{\mathrm{ood}}$. The ratio $\rho$ determines
the fraction of training prompt groups drawn from GraftSet-Video. Each such
group combines $n_{\mathrm{ood}}$ of the prompt's four stored exemplars with
$G-n_{\mathrm{ood}}$ fresh online rollouts. The remaining groups use original
training prompts with $G$ online rollouts each and no external exemplar supervision.

\begin{table}[!t]
    \centering
    \caption{HPSv3 optimization of Wan2.1-1.3B with CFG-enabled evaluation.
    OOD methods use $\rho=0.167$ and $n_{\mathrm{ood}}=4$.
    All metrics are higher-is-better (definitions in
    Appendix~\ref{app:training_details}). Bold marks column bests among
    FlashOOD, NFTOOD, and GraftOOD only.}
    \label{tab:hpsv3_main}
    \setlength{\tabcolsep}{3pt}
    \begin{tabular*}{\textwidth}{@{\extracolsep{\fill}}l*{4}{c}|*{6}{c}@{}}
        \toprule
        Method & AQ & IQ & SC & OC & DD & VR & VQ & MQ & TA & HPSv3 \\
        \midrule
        Wan2.1 T2V 1.3B & 65.63 & 66.79 & 94.58 & 88.70 & 48.61
                        & 1.26 & -0.59 & -0.34 & -0.66 & 4.70 \\
        \midrule
        Flash-GRPO & 67.46 & 68.10 & 95.85 & 90.36 & 28.67
                   & 1.65 & -0.69 & -0.43 & -0.59 & 5.11 \\
        DiffusionNFT & 68.30 & 71.50 & 97.49 & 90.9
                     & 14.24 & 5.20 & -0.09 & 0.21
                     & -0.51 & 7.23 \\
        FlowAWR & 69.18 & 71.01 & 98.48 & 92.53 & 4.86
                & 5.11 & 0.06 & 0.47 & -0.65
                & 8.10 \\
        \midrule
        FlashOOD & 65.87 & 67.91 & 94.68 & 88.29
                 & \textbf{58.26} & 1.65 & -0.62 & -0.47
                 & -0.50 & 5.31 \\
        NFTOOD & 66.10 & 70.52 & 96.99
               & 87.91 & 41.12 & 3.90
               & -0.45 & -0.25 & -0.47 & 6.55 \\
        GraftOOD & \textbf{67.21} & \textbf{71.26} & \textbf{98.20}
                 & \textbf{90.10} & 44.41 & \textbf{5.08}
                 & \textbf{-0.25} & \textbf{-0.07} & \textbf{-0.45}
                 & \textbf{7.32} \\
        \bottomrule
    \end{tabular*}
    \vspace{-0.1in}
\end{table}

\begin{figure}[!t]
    \centering
    \includegraphics[width=\linewidth]{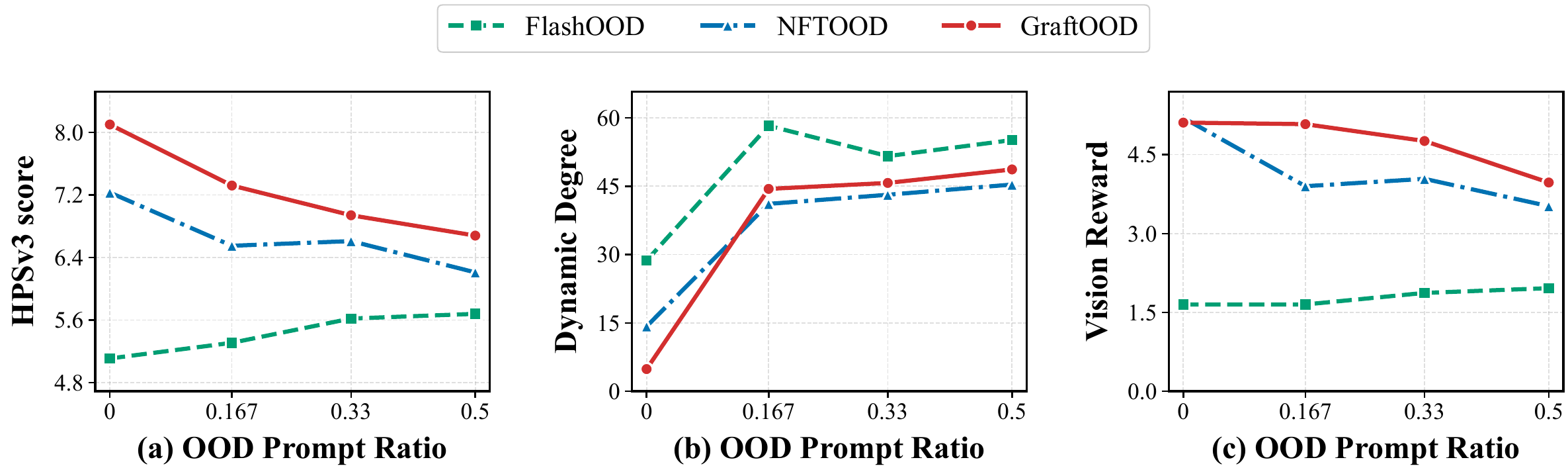}
    \vspace{-0.2in}
    \caption{\textbf{OOD prompt-ratio sweep.}
    \textbf{(a)} HPSv3, \textbf{(b)} DD, and \textbf{(c)} VisionReward with
    $n_{\mathrm{ood}}=4$ at $\rho>0$. At $\rho=0$, FlashOOD, NFTOOD, and
    GraftOOD use Flash-GRPO, DiffusionNFT, and FlowAWR, respectively,
    without introducing external exemplars.}
    \label{fig:ratio_comparison}
    \vspace{-0.08in}
\end{figure}

\section{Experiments}
\label{sec:experiments}

\subsection{Experimental Setup}
\label{subsec:experimental_setup}

\noindent\textbf{Datasets, Models, and Baselines.}
Following DanceGRPO~\citep{xue2025dancegrpo}, we use 19,700 training prompts
and add exemplars from GraftSet-Video. On Wan2.1-1.3B~\citep{wan2025}, we
compare the base model, Flash-GRPO~\citep{he2026flash},
DiffusionNFT~\citep{zheng2025diffusionnft}, and FlowAWR~\citep{fu2026flowawr},
alongside their exemplar-augmented variants FlashOOD, NFTOOD, and GraftOOD,
as defined by the three training objectives in Section~\ref{subsec:group_grafting}.

\noindent\textbf{Implementation Details.}
Training uses 20-step rollouts, classifier-free guidance (CFG) scale 4.5,
and group size $G=8$. NFT/AWR variants use EMA sampling and CFG-free updates.
We train rank-32 attention LoRA adapters with AdamW at learning
rate $10^{-4}$, freezing the backbone. The main OOD comparison uses
$\rho=0.167$ and $n_{\mathrm{ood}}=4$. All methods use 50 sampling steps and
CFG scale 4.5 for evaluation, generating 81-frame videos at $832\times480$
and 16 fps. Further settings, including LoRA scaling, numerical precision,
and regularization, appear in Appendix~\ref{app:training_details}.

\noindent\textbf{Rewards and Evaluation.}
We optimize HPSv3~\citep{ma2025hpsv3}, averaging the top 30\% of per-frame
scores following LongCat-Video~\citep{team2025longcat}. We report VBench's
Aesthetic Quality, Imaging Quality, Subject Consistency, and Object Class
on its official prompt set~\citep{huang2024vbench}. On 300 VidProM prompts,
we measure Dynamic Degree, VisionReward~\citep{xu2026visionreward},
VideoAlign's Video Quality, Motion Quality, and Text
Alignment~\citep{liu2026improving}, and HPSv3.
Appendix~\ref{app:training_details} details the prompt sets, score aggregation,
and evaluation protocol.

\begin{figure}[!t]
    \centering
    \includegraphics[width=\linewidth]{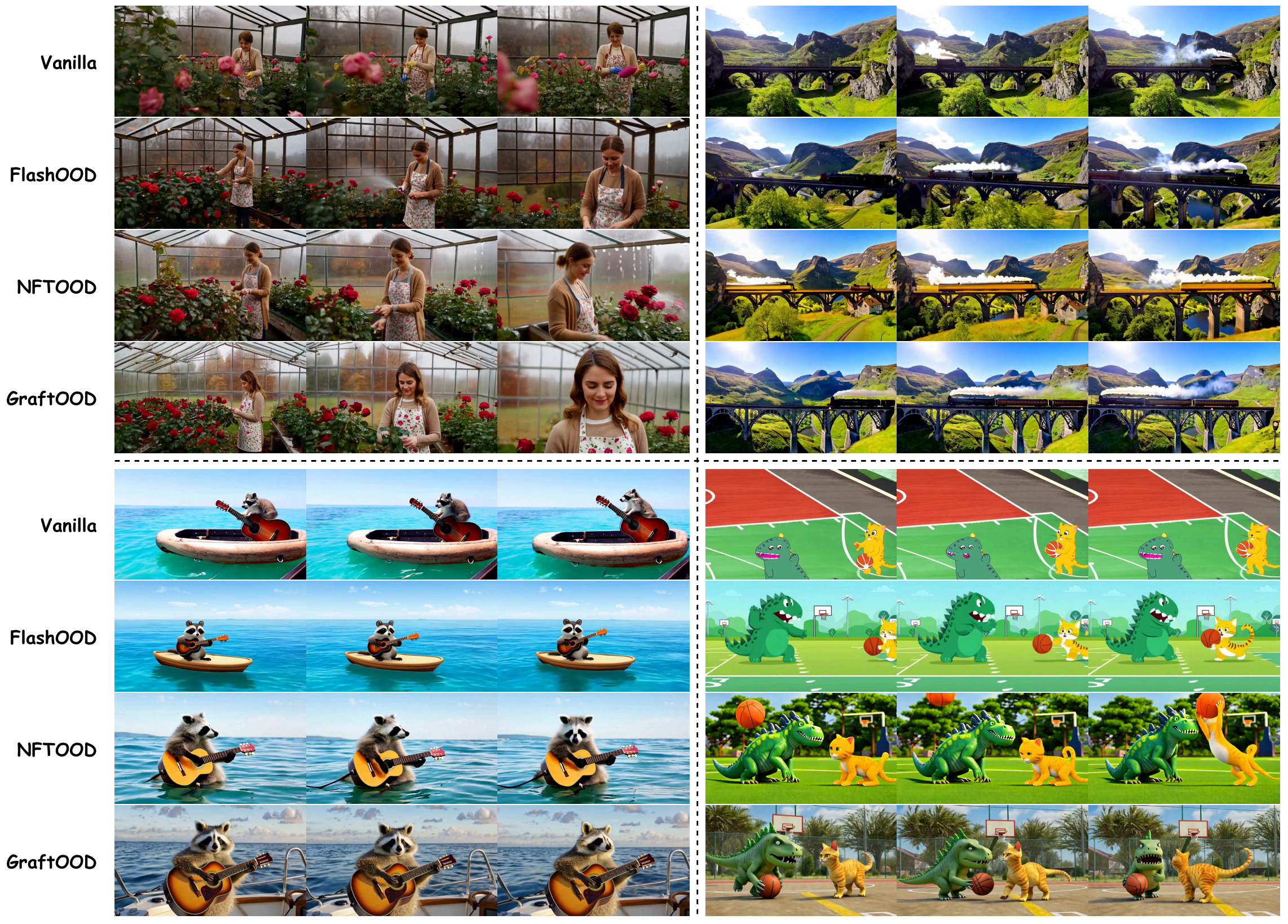}
    \vspace{-0.2in}
    \caption{\textbf{Qualitative comparison.} Each prompt block shows Vanilla,
    FlashOOD, NFTOOD, and GraftOOD from top to bottom, with three frames
    per method in temporal order.}
    \label{fig:graftood_compare}
    \vspace{-0.08in}
\end{figure}

\subsection{Main Results}
\label{subsec:main_comparison}

Relative to FlowAWR, GraftOOD raises Dynamic Degree (DD) from 4.86 to 44.41,
while HPSv3 changes from 8.10 to 7.32 and VisionReward from 5.11 to 5.08
(Table~\ref{tab:hpsv3_main}). Although DD remains below the base model's
48.61, HPSv3 and VisionReward exceed its 4.70 and 1.26.
GraftOOD thus retains more dynamics than reward-only FlowAWR while preserving
reward and visual-score gains over the base model, despite the remaining
gap in measured dynamics. All three exemplar-augmented methods raise DD
over their respective online baselines, but HPSv3 rises only for FlashOOD
and falls for NFTOOD and GraftOOD under the shared main-comparison configuration.

GraftOOD leads nine of ten metrics among OOD methods, including all four
VBench metrics. FlashOOD attains the highest DD (58.26), but the bottom-right
basketball example in Figure~\ref{fig:graftood_compare} shows character
deformation. DD uses RAFT-estimated optical-flow magnitudes
\citep{huang2024vbench,teed2020raft} without distinguishing coherent motion
from deformation. Deformation can therefore inflate DD, offering a possible
explanation for FlashOOD's score rather than evidence of better motion.
GraftOOD exceeds both OOD baselines on SC, VQ, and MQ at this setting.

\subsection{Ablation Studies}
\label{subsec:exemplar_count_ablation}

\paragraph{OOD prompt ratio at a fixed exemplar count.}
At $n_{\mathrm{ood}}=4$, increasing $\rho$ from 0.167 to 0.5 raises GraftOOD's
DD from 44.41 to 48.68 but lowers HPSv3 and VisionReward
(Figure~\ref{fig:ratio_comparison}). GraftOOD still leads both reward metrics
among OOD methods at every tested positive ratio. The sweep therefore
shows a dynamics--reward trade-off, not a uniform improvement.

\begin{figure}[!t]
    \centering
    \includegraphics[width=\linewidth]{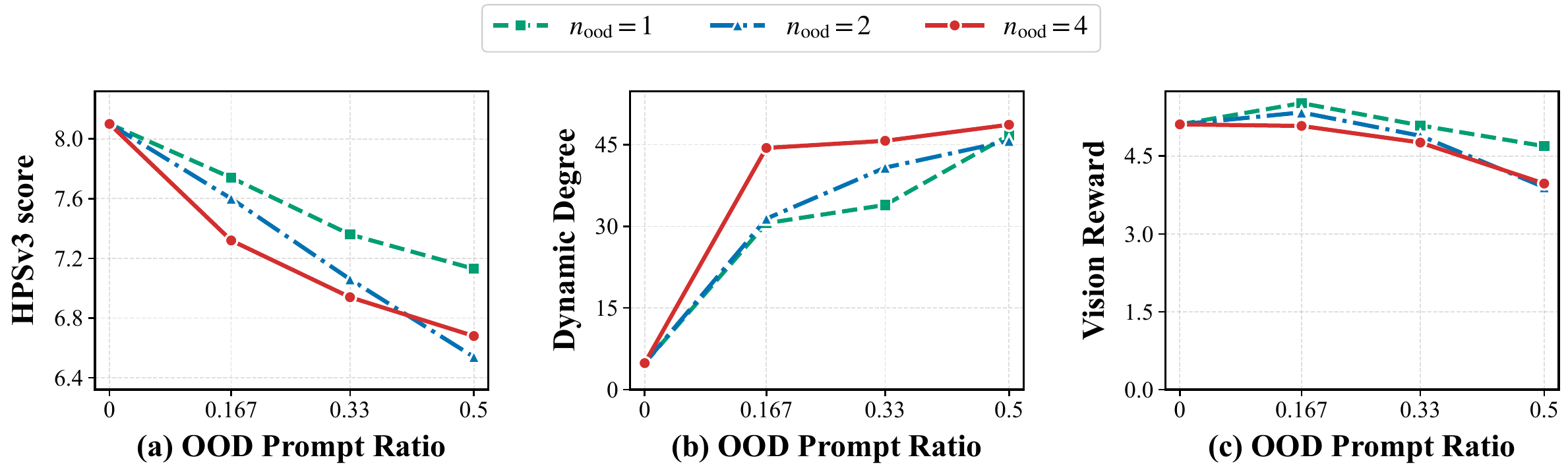}
    \vspace{-12pt}
    \caption{\textbf{GraftOOD exemplar-count ablation.}
    \textbf{(a)} HPSv3, \textbf{(b)} DD, and \textbf{(c)} VisionReward versus
    $\rho$ for $n_{\mathrm{ood}}\in\{1,2,4\}$ ($G=8$). Curves share one
    FlowAWR baseline at $\rho=0$.}
    \label{fig:count_ablation}
    \vspace{-0.08in}
\end{figure}

\paragraph{Exemplar count at a fixed OOD prompt ratio.}
Among $n_{\mathrm{ood}}\in\{1,2,4\}$, one exemplar yields the highest HPSv3
and VisionReward at every tested positive $\rho$, while four yield the
highest DD (Figure~\ref{fig:count_ablation}). At $\rho=0.167$, reducing the
count from four to two lowers DD from 44.41 to 31.41, but raises HPSv3
from 7.32 to 7.60. Two exemplars score below both alternatives on all three
metrics at $\rho=0.5$. Since $G=8$ is fixed,
exemplar count also changes the online sample count for reward normalization:
this tests group composition, not exemplar supervision alone.
Appendix~\ref{app:additional_results} reports the full ratio and exemplar-count
sweeps with all evaluated quality and reward metrics.

\FloatBarrier

\section{Related Work}
\label{sec:related_work}

\paragraph{Reward Alignment for Video Generation.}
Preference alignment for video generation includes pairwise objectives,
such as VideoDPO~\citep{liu2024videodpo}, and online reward optimization.
DDPO and DPOK formulate diffusion fine-tuning through policy
gradients~\citep{black2024ddpo,fan2023dpok}. Flow-GRPO and DanceGRPO extend
group-relative optimization to flow-based visual generation using stochastic
sampling~\citep{liu2026flow,xue2025dancegrpo}, while MixGRPO and Flash-GRPO
adopt mixed ODE--SDE sampling and single-timestep updates,
respectively~\citep{li2025mixgrpo,he2026flash}. Alternatively, DiffusionNFT
optimizes reward-weighted regression on the forward
process~\citep{zheng2025diffusionnft}, and FlowAWR derives advantage-weighted
velocity rectification~\citep{fu2026flowawr}. GraftOOD builds on this
regression formulation but addresses a different question: how external
video exemplars can supplement reward optimization when the proxy
inadequately constrains temporal behavior of the generated video sequences.

\paragraph{External Data for Reinforcement Learning.}
External experience has been incorporated into online RL through different
learning signals. RLPD mixes offline transitions with online replay for
actor--critic learning~\citep{ball2023efficient}, while LUFFY incorporates
expert reasoning trajectories into mixed-policy group-relative
optimization~\citep{yan2026learning}. GraftOOD addresses a different
supervision problem: external videos specify desired sequence-level
behavior but provide no generating denoising trajectories. This distinction
concerns the form of external supervision, not merely its source: each
video provides a regression target without a recorded sequence of policy
actions. We use these terminal exemplars to construct supervised velocity targets and combine
them with reward-relative online targets within prompt groups, excluding
exemplars from online reward normalization. Beyond combining data sources,
our analysis characterizes the resulting exemplar-guided velocity field.
This formulation supplements attributes inadequately constrained by the
proxy reward without an additional motion reward or training-time
reward-model scoring of external exemplars.

\section{Conclusion}
\label{sec:conclusion}

We identified motion collapse in video reward optimization, where increasing
proxy rewards can coincide with declining video dynamics. We proposed
GraftOOD to incorporate external sequence-level supervision into
group-relative velocity regression without an additional supervised-loss
coefficient. Our analysis characterizes its pointwise optimum as a
source-posterior-weighted combination of online and exemplar target fields.
With GraftSet-Video, experiments on Wan2.1-1.3B show higher measured dynamics
than reward-only FlowAWR, while exemplar-frequency and count ablations
reveal trade-offs with HPSv3 and VisionReward. These findings support
external exemplars as a complement to scalar reward feedback, rather than
a replacement for reward optimization. Exemplar selection and motion-quality
evaluation remain limitations: higher Dynamic Degree alone does not
establish coherent motion or improved temporal quality in generated videos.

\clearpage
\subsection*{AI use statement}
We used generative AI tools to assist with manuscript drafting and revision,
including wording, LaTeX formatting. AI-assisted checks do
not replace author verification. The authors take responsibility for the
manuscript, including its technical claims, derivations, experimental
results, and references.

\subsection*{Ethics statement}
GraftOOD uses pretrained video models, learned reward functions, and
generated video exemplars. It may inherit biases from these sources,
and higher reward scores do not by themselves ensure safe or unbiased
outputs. We have not identified additional ethical concerns specific
to this study beyond these general risks.

\subsection*{Reproducibility statement}
We describe the sample-averaged training objective and sampling assumptions
in Appendix~\ref{app:dual_source_analysis}, with the proof of
Proposition~\ref{prop:convex} in Appendix~\ref{app:proof_convex}.
GraftSet-Video construction and source statistics are described in
Section~\ref{subsec:ood_corpus} and Appendix~\ref{app:corpus_details}.
We will publicly release GraftSet-Video upon acceptance of the paper
to support reproducibility and further research on exemplar-guided
video reward optimization.
Training configurations, reward scaling, exemplar filtering, and evaluation
protocols are provided in Appendix~\ref{app:training_details}.
The motion-quality reward study, additional method comparisons, and
exemplar-ratio and count ablations are reported in
Appendix~\ref{app:additional_results}.
Additional qualitative examples and their prompts appear in
Appendix~\ref{app:more_qualitative}.

\bibliography{iclr2027_conference}
\bibliographystyle{iclr2027_conference}

\clearpage
\appendix
\section*{Appendix Contents}
\pdfbookmark[0]{Appendix Contents}{appendix-contents}
\begingroup
\newcommand{\appendixentry}[2]{%
    \noindent\hspace*{#1}\hyperref[#2]{\ref*{#2}\quad\nameref*{#2}}%
    \nobreak\dotfill\pageref{#2}\par
}
\appendixentry{0pt}{app:dual_source_analysis}
\appendixentry{1em}{app:sample_averaged}
\appendixentry{1em}{app:proof_convex}
\appendixentry{0pt}{app:ratio_comparisons}
\appendixentry{1em}{app:corpus_details}
\appendixentry{1em}{app:training_details}
\appendixentry{1em}{app:additional_results}
\appendixentry{0pt}{app:limitations}
\appendixentry{0pt}{app:more_qualitative}
\appendixentry{1em}{app:qualitative_comparisons}
\appendixentry{1em}{app:qualitative_prompts}
\endgroup

\section{Analysis of Dual-Source Regression}
\label{app:dual_source_analysis}

\subsection{Sample-Averaged Implementation}
\label{app:sample_averaged}
An original training prompt uses $G$ rollouts; an exemplar prompt uses
$n_{\mathrm{ood}}$ stored videos and $M=G-n_{\mathrm{ood}}\ge2$ fresh
rollouts. All members share the prompt and timestep $t_s$, with independent
fresh noise. Let $\mathcal{I}_s^{\mathrm{roll}}$ index the online subset,
of size $m_s\in\{G,M\}$. Only rollouts are scored during training.
For $R_i=R(c_s,x_1^{(i)})$ and $\gamma>0$, define
\begin{equation}
    \phi_s=\frac{1}{m_s}\sum_{k\in\mathcal{I}_s^{\mathrm{roll}}}
    \exp(R_k/\gamma),
    \qquad
    a_s^{(i)}=
    \begin{cases}
        \widehat{\Adv}_{0,s}^{(i)}=\exp(R_i/\gamma)/\phi_s-1,
        & i\in\mathcal{I}_s^{\mathrm{roll}},\\
        1, & \text{otherwise}.
    \end{cases}
    \label{eq:graft_online_advantage}
\end{equation}
Exemplars do not enter $\phi_s$; rollout advantages sum to zero within their
subset. A single rollout would therefore have zero centered advantage.
Write $v_\theta^{(i)}=v_\theta(x_{t_s}^{(i)},c_s,t_s)$, likewise for
$v^{\mathrm{old},(i)}$, and $u_{t_s}^{(i)}=x_1^{(i)}-x_0^{(i)}$.
The group-averaged loss is
\begin{equation}
    \widehat{\mathcal{L}}_{\mathrm{GraftOOD}}(\theta)
    =\frac{1}{G}\sum_{i=1}^{G}
    \left\|v_\theta^{(i)}-
    \left[v^{\mathrm{old},(i)}+a_s^{(i)}
    \left(u_{t_s}^{(i)}-v^{\mathrm{old},(i)}\right)\right]\right\|_2^2.
    \label{eq:graftood_loss}
\end{equation}
For an exemplar, $a_s^{(i)}=1$ cancels the reference field pointwise,
so using $v^{\mathrm{old}}$ here gives the same target $u_t$ as the
source-specific expression in \Eqref{eq:graftood_source_objective}.
Targets are detached. For an exemplar prompt, the expected group average
is \Eqref{eq:graftood_source_objective}; original prompts retain only the
online term.

For a fixed temperature, independent re-noising makes $x_0$ independent of
the terminal given the prompt. The group normalizer estimates the noise-end denominator
$\phi(x_0,c,0)$ from \Eqref{eq:awr_population_advantage}. Reusing
$\widehat{\Adv}_0$ at other timesteps approximates, rather than computes,
their state-conditional coefficients. The subscript zero denotes conditioning
time; rewards are evaluated on clean terminals.

\subsection{Sampling Assumptions and Proof}
\label{app:proof_convex}

\paragraph{Sampling model and assumptions.}
Fix $c_s$, the reference parameters, and the group-sampling law for one
optimization round. Draw a group with $M$ online rollouts and
$n_{\mathrm{ood}}$ exemplars, then select one member uniformly. Its source
$S\in\{\mathrm{on},\mathrm{ood}\}$ has prior
$\Pr(S=\mathrm{ood}\mid c_s)=\delta=n_{\mathrm{ood}}/G$. The source-conditional
terminal laws are $p^{\mathrm{old}}(\cdot\mid c_s)$ and
$p^{\mathrm{ood}}(\cdot\mid c_s)$, with the latter reflecting the exemplar
selection rule. Draw the shared $t_s\sim\nu$ independently of the collected
group and re-noise each member independently. Linearity of expectation makes
this shared draw equivalent in expected loss to independent per-member draws
from $\nu$, though their variances may differ. All almost-everywhere statements
below refer to this sampling law, including when $\nu$ is discrete.

For the selected member, set $Z=(c_s,x_t,t)$ with $t=t_s$ and
$Y=v^{\mathrm{old}}(Z)+a(u_t-v^{\mathrm{old}}(Z))$, using
\Eqref{eq:graft_online_advantage}. Assume $\mathbb{E}\|Y\|_2^2<\infty$
and that the two noised source laws admit densities with respect to a common
dominating measure for $\nu$-almost every $t$. Uniform member selection
reproduces \Eqref{eq:graftood_source_objective}. Wherever the mixture density
is positive, Bayes' rule gives
$\Pr(S=\mathrm{ood}\mid Z)=w_s(x_t,t)$ in
\Eqref{eq:prompt_source_posterior}. Conditional means need only be defined
on each source's support; a zero-posterior term is omitted.

\paragraph{Proof of Proposition~\ref*{prop:convex}.}
Under the fixed-prompt sampling law above, the conditional variance
decomposition gives, for any measurable field $v$ of finite risk,
\begin{equation}
    \mathbb{E}\|v(Z)-Y\|_2^2
    =\mathbb{E}\|v(Z)-\mathbb{E}[Y\mid Z]\|_2^2
    +\mathbb{E}\|Y-\mathbb{E}[Y\mid Z]\|_2^2.
    \label{eq:dual_variance_decomposition}
\end{equation}
The second term is independent of $v$, so the minimizer is
$v^\dagger(Z)=\mathbb{E}[Y\mid Z]$, unique up to sets of training-state
measure zero. Define
$v_{\mathrm{old}}^\star(Z)=\mathbb{E}[Y\mid Z,S=\mathrm{on}]$.
For exemplars, $Y=u_t$ and conditional flow matching gives
$\mathbb{E}[Y\mid Z,S=\mathrm{ood}]=v^{\mathrm{ood}}(Z)$.
Conditioning on the source therefore yields
\begin{equation*}
    v^\dagger(Z)=(1-w_s)v_{\mathrm{old}}^\star(Z)
    +w_s v^{\mathrm{ood}}(Z),
\end{equation*}
which is \Eqref{eq:pointwise_optimum}. The online conditional mean includes
the other rollout members that determine the finite-group coefficient and
the rollout-batch statistics used to set $\gamma$;
no population-advantage substitution is needed.

This is a characterization over measurable fields. The empirical online
target need not equal the exact KL-optimal field, and zero-sum group
advantages do not cancel reference fields evaluated at different states.
The result does not establish neural-network attainability, training
convergence, terminal-distribution mixing, or video-quality improvement.

\section{Experimental Details and Additional Results}
\label{app:ratio_comparisons}

\subsection{GraftSet-Video Details}
\label{app:corpus_details}

\begin{table}[!ht]
    \begin{minipage}[t]{0.58\linewidth}
        \vspace{0pt}
        \noindent\textbf{Source counts and selection.}
        Table~\ref{tab:ood_sources} gives exact source counts, supplementing the
        proportions and mean HPSv3 scores in Figure~\ref{fig:graftset_sources}.
        Construction and training-time allocation follow
        Section~\ref{subsec:ood_corpus}: supplementary outputs fill unoccupied
        exemplar slots rather than replacing retained videos through a pooled
        reranking of collected and supplementary candidates.

        \smallskip
        \noindent\textbf{Corpus statistics.}
        HPSv3 scores span $[-0.139,15.131]$, with mean $9.735$ and standard
        deviation $1.413$. Of the exemplar prompts, 52.0\% contain Chinese text,
        and the mean prompt length is 388 characters. Video format and prompt-set
        separation are specified in Section~\ref{subsec:ood_corpus}.
    \end{minipage}\hfill
    \begin{minipage}[t]{0.38\linewidth}
        \vspace{0pt}
        \centering
        \setlength{\abovecaptionskip}{0pt}
        \caption{Exemplars by source.}
        \label{tab:ood_sources}
        \begin{tabular}{lr}
            \toprule
            System & Videos \\
            \midrule
            \texttt{wan2\_2}        & 2,164 \\
            \texttt{vidu\_q2}       & 1,163 \\
            \texttt{seedance\_1\_5} & 1,144 \\
            \texttt{kling\_2\_5}    & 1,098 \\
            \texttt{kling3}         &   915 \\
            \texttt{seedance2}      &   911 \\
            \texttt{veo3\_1}        &   883 \\
            \texttt{hailuo}         &   830 \\
            \texttt{sora2}          &   480 \\
            \midrule
            Total                  & 9,588 \\
            \bottomrule
        \end{tabular}
    \end{minipage}
\end{table}
\FloatBarrier

\subsection{Experimental Setup}
\label{app:training_details}

\paragraph{Training and sampling configurations.}

All experiments use Wan2.1-1.3B \citep{wan2025} and 19,700 training prompts
following DanceGRPO \citep{xue2025dancegrpo}, with additional exemplar prompts
and videos from GraftSet-Video (Section~\ref{subsec:ood_corpus}). Videos
contain 81 frames at $832\times480$ resolution and 16 frames per second.
Online sampling uses classifier-free guidance (CFG); NFT/AWR variants
disable CFG for training updates. All reported evaluations, including the base
model, use 50 sampling steps and a CFG scale of 4.5. Training rollouts use
20 sampling steps, a CFG scale of 4.5, and group size $G=8$; NFT/AWR variants
use EMA reference policies for sampling.
We train rank-32 LoRA adapters with scaling factor 64 on the attention
projections, keeping the backbone frozen. We use bfloat16 and AdamW with
learning rate $10^{-4}$, weight decay $10^{-4}$, and gradient norm clipping
at 1.0. FlowAWR and GraftOOD set $\gamma$ to the global standard deviation
of online-rollout rewards in the current batch, shared across prompt groups
and excluding exemplars. FlashOOD uses $\lambda_{\mathrm{ood}}=1$ in all
configurations. GraftOOD shares a regression timestep within each prompt
group and independently re-noises each video.

The main comparison uses $\rho=0.167$ and $n_{\mathrm{ood}}=4$.
The ablations separate two controls: the frequency of exemplar prompt
groups, $\rho$, and their within-group exemplar count, $n_{\mathrm{ood}}$.
The experimental grid uses $\rho\in\{0.167,0.33,0.5\}$ and
$n_{\mathrm{ood}}\in\{4,2,1\}$ with $G=8$, retaining four, six, or seven
online rollouts per exemplar group, respectively. All configurations use
the same GraftSet-Video candidate pool, with four stored videos per prompt.
Reducing $n_{\mathrm{ood}}$ changes how many candidates are used in a group,
not the corpus or prompt population.

\paragraph{Reward and evaluation protocols.}

The main experiments optimize HPSv3 \citep{ma2025hpsv3}, averaging the top
30\% of per-frame scores following LongCat-Video \citep{team2025longcat}.
For offline corpus construction, the initial selection from public resources
retains videos with HPSv3 scores above 7, before supplementary generation
fills the remaining exemplar slots. This initial threshold does not impose
a score floor on the supplemented corpus. External exemplars do not enter
training-time reward normalization.
We report VBench's Aesthetic Quality (AQ), Imaging Quality (IQ), Subject
Consistency (SC), and Object Class (OC) on its official prompt set
\citep{huang2024vbench}. On the 300-prompt VidProM evaluation set, we measure
Dynamic Degree (DD), VisionReward (VR) \citep{xu2026visionreward},
VideoAlign's Video Quality (VQ), Motion Quality (MQ), and Text Alignment (TA)
\citep{liu2026improving}, and HPSv3. Motion Quality serves as an evaluation
metric here; its use as a training reward in the motivation study is
discussed in Section~\ref{subsec:dual_source}.

These evaluation sets remain unchanged across prompt ratios and exemplar
counts. DD measures detected dynamics rather than motion plausibility and
should be read alongside the other metrics and qualitative examples in
Section~\ref{subsec:main_comparison}.

\subsection{Additional Quantitative Results}
\label{app:additional_results}

\paragraph{Motion-quality reward optimization.}

\begin{figure}[!ht]
    \centering
    \begin{minipage}{0.49\linewidth}
        \centering
        \includegraphics[width=\linewidth,trim=0 45 0 0,clip]{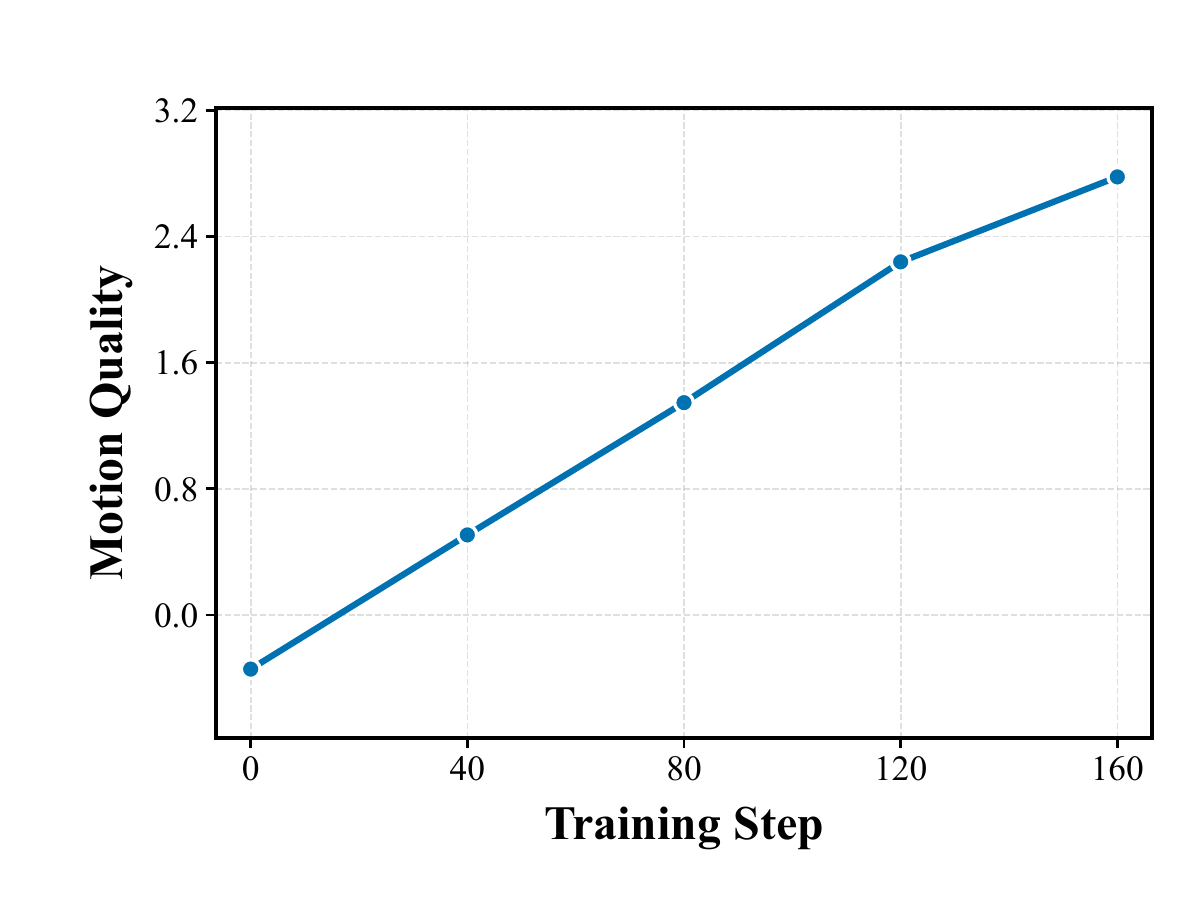}
        \par\small (a) Training step
    \end{minipage}\hfill
    \begin{minipage}{0.49\linewidth}
        \centering
        \includegraphics[width=\linewidth,trim=0 45 0 0,clip]{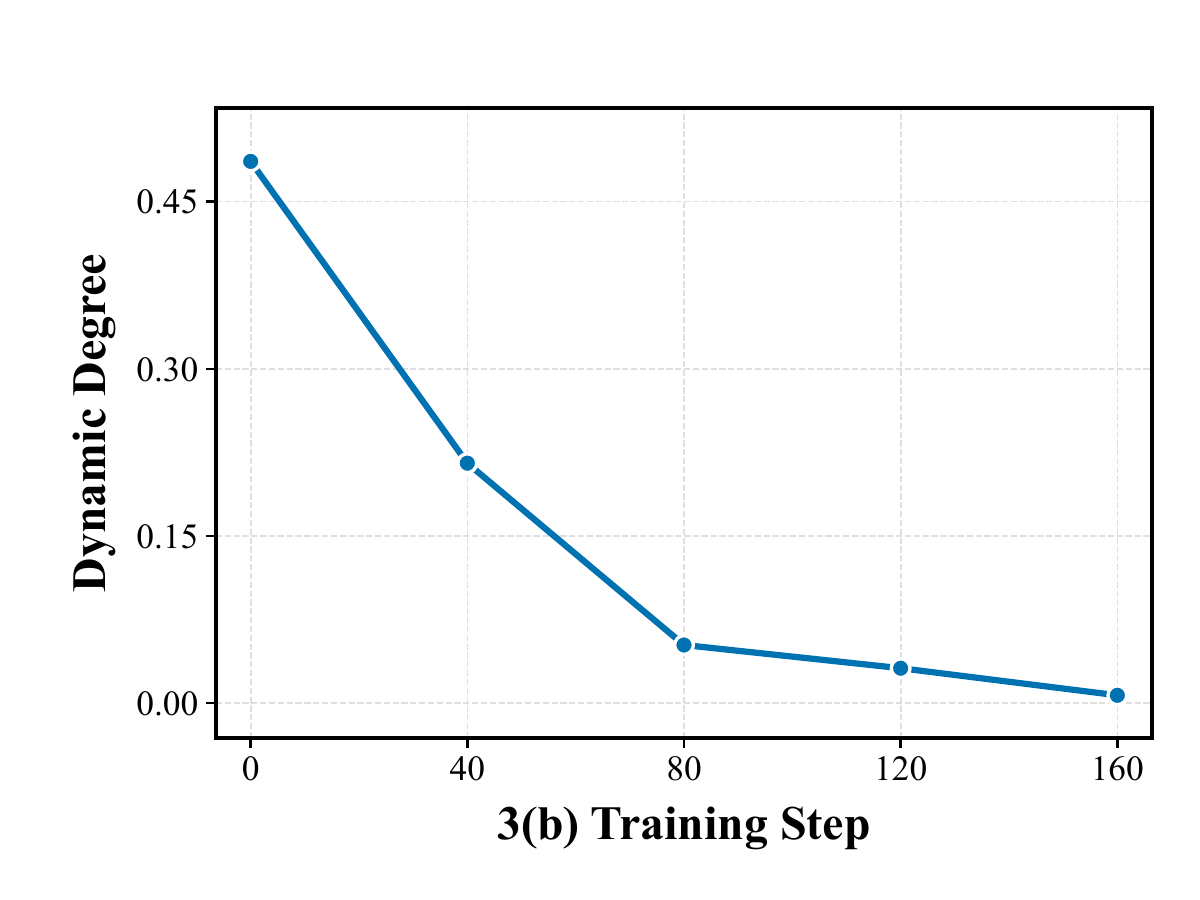}
        \par\small (b) Training step
    \end{minipage}
    \caption{\textbf{Motion-quality reward and video dynamics.}
    VideoAlign Motion Quality reward (left) and Dynamic Degree (right)
    during the same reward-optimization run.}
    \label{fig:mq_reward_dynamics}
\end{figure}

Figure~\ref{fig:mq_reward_dynamics} supplements the motivation study in
Section~\ref{subsec:dual_source}. During optimization of VideoAlign Motion
Quality, its reward increases while Dynamic Degree decreases. This example
shows that using a video-level motion-quality reward does not by itself
prevent the loss of measured dynamics.

\paragraph{Method comparisons at four exemplars.}

Tables~\ref{tab:ratio_0167}--\ref{tab:ratio_05} compare FlashOOD, NFTOOD, and
GraftOOD at each ratio with $n_{\mathrm{ood}}=4$. Bold values denote the
best result among these three methods in the same column and table, not
across ratios. The $\rho=0.167$ results repeat the OOD rows of
Table~\ref{tab:hpsv3_main}; the three tables provide the positive-ratio
points in Figure~\ref{fig:ratio_comparison}. The $\rho=0$ points in that
figure instead use each method's reward-only baseline from the main table.
GraftOOD has the highest HPSv3 and VR at all three positive ratios, while
FlashOOD has the highest DD. Within GraftOOD, increasing $\rho$ raises DD
but lowers HPSv3 and VR over the reported range.

\begin{table*}[!ht]
    \centering
    \caption{OOD comparison at $\rho=0.167$, $n_{\mathrm{ood}}=4$.
    Bold marks column bests.}
    \label{tab:ratio_0167}
    \setlength{\tabcolsep}{3pt}
    \begin{tabular*}{\textwidth}{@{\extracolsep{\fill}}l*{4}{c}|*{6}{c}@{}}
        \toprule
        Method & AQ & IQ & SC & OC & DD & VR & VQ & MQ & TA & HPSv3 \\
        \midrule
        FlashOOD & 65.87 & 67.91 & 94.68 & 88.29
                 & \textbf{58.26} & 1.65 & -0.62
                 & -0.47 & -0.50 & 5.31 \\
        NFTOOD & 66.10 & 70.52 & 96.99
               & 87.91 & 41.12 & 3.90
               & -0.45 & -0.25 & -0.47 & 6.55 \\
        GraftOOD & \textbf{67.21} & \textbf{71.26} & \textbf{98.20}
                 & \textbf{90.10} & 44.41 & \textbf{5.08}
                 & \textbf{-0.25} & \textbf{-0.07} & \textbf{-0.45}
                 & \textbf{7.32} \\
        \bottomrule
    \end{tabular*}
    \vspace{-4pt}
\end{table*}

\begin{table*}[!ht]
    \centering
    \caption{OOD comparison at $\rho=0.33$, $n_{\mathrm{ood}}=4$.
    Bold marks column bests.}
    \label{tab:ratio_033}
    \setlength{\tabcolsep}{3pt}
    \begin{tabular*}{\textwidth}{@{\extracolsep{\fill}}l*{4}{c}|*{6}{c}@{}}
        \toprule
        Method & AQ & IQ & SC & OC & DD & VR & VQ & MQ & TA & HPSv3 \\
        \midrule
        FlashOOD & 65.78 & 67.17 & 95.05 & 88.21
                         & \textbf{51.60} & 1.87 & -0.52 & -0.38
                         & -0.49 & 5.62 \\
        NFTOOD & \textbf{66.41} & 69.11 & 96.95
                           & 89.85 & 43.12 & 4.04 & -0.38
                           & -0.21 & \textbf{-0.46} & 6.61 \\
        GraftOOD & 66.34 & \textbf{71.39} & \textbf{97.04}
                  & \textbf{90.19} & 45.72 & \textbf{4.76}
                  & \textbf{-0.32} & \textbf{-0.14} & -0.50
                  & \textbf{6.94} \\
        \bottomrule
    \end{tabular*}
    \vspace{-4pt}
\end{table*}

\begin{table*}[!ht]
    \centering
    \caption{OOD comparison at $\rho=0.5$, $n_{\mathrm{ood}}=4$.
    Bold marks column bests.}
    \label{tab:ratio_05}
    \setlength{\tabcolsep}{3pt}
    \begin{tabular*}{\textwidth}{@{\extracolsep{\fill}}l*{4}{c}|*{6}{c}@{}}
        \toprule
        Method & AQ & IQ & SC & OC & DD & VR & VQ & MQ & TA & HPSv3 \\
        \midrule
        FlashOOD & 66.10 & 68.40 & 94.72 & 87.10
                         & \textbf{55.13} & 1.96 & -0.63 & -0.49
                         & -0.51 & 5.68 \\
        NFTOOD & \textbf{66.27} & 69.01 & 95.77
                           & \textbf{89.88} & 45.36 & 3.52
                           & \textbf{-0.37} & \textbf{-0.14}
                           & -0.46 & 6.21 \\
        GraftOOD & 66.03 & \textbf{69.21} & \textbf{96.75} & 88.92
                  & 48.68 & \textbf{3.97} & -0.43 & -0.25
                  & \textbf{-0.39} & \textbf{6.68} \\
        \bottomrule
    \end{tabular*}
    \vspace{-4pt}
\end{table*}

\paragraph{GraftOOD ratio and count ablations.}

Tables~\ref{tab:graftood_n4}--\ref{tab:graftood_n1} arrange the GraftOOD
results by exemplar count. All three tables include the same $\rho=0$
FlowAWR baseline from Table~\ref{tab:hpsv3_main}. At zero ratio, no exemplar
groups are sampled and $n_{\mathrm{ood}}$ is inactive; these entries repeat
one baseline rather than represent separate count-specific runs. Reading
down a table varies the frequency of exemplar groups while keeping their
composition fixed at positive ratios. Comparing the same positive $\rho$
across tables varies the number of exemplars within each such group while
keeping its sampling frequency fixed. The positive-ratio rows of
Table~\ref{tab:graftood_n4} collect the GraftOOD results from
Tables~\ref{tab:ratio_0167}--\ref{tab:ratio_05}; they do not represent
additional runs. Tables~\ref{tab:graftood_n2}
and~\ref{tab:graftood_n1} report the corresponding results for smaller
counts. Figure~\ref{fig:count_ablation} compares HPSv3, DD, and VisionReward
across these settings. Bold values mark column bests among positive ratios
within each fixed-count table; the shared $\rho=0$ baseline is excluded.

Under the group sampling scheme in Section~\ref{subsec:ood_corpus}, the
fraction of training samples supplied by exemplars is
$\rho n_{\mathrm{ood}}/8$. Thus, increasing either control increases exemplar
exposure, but the two interventions differ: $\rho$ changes the frequency of
exemplar prompts, whereas $n_{\mathrm{ood}}$ also changes the number of online
samples within those groups. The latter affects the sample set used for
online reward normalization. The two comparisons therefore assess distinct
training configurations rather than interchangeable settings of a single
loss weight.

\begingroup
\makeatletter
\setlength{\@fptop}{0pt}
\setlength{\@fpsep}{12pt}
\setlength{\@fpbot}{0pt plus 1fil}
\makeatother
\begin{table*}[!ht]
    \centering
    \caption{GraftOOD ratio sweep with $n_{\mathrm{ood}}=4$ ($G=8$).
    The $\rho=0$ row repeats FlowAWR; positive rows repeat
    Tables~\ref{tab:ratio_0167}--\ref{tab:ratio_05}.
    Bold marks column bests among positive ratios only.}
    \label{tab:graftood_n4}
    \setlength{\tabcolsep}{3pt}
    \begin{tabular*}{\textwidth}{@{\extracolsep{\fill}}l*{4}{c}|*{6}{c}@{}}
        \toprule
        $\rho$ & AQ & IQ & SC & OC & DD & VR & VQ & MQ & TA & HPSv3 \\
        \midrule
        0 & 69.18 & 71.01 & 98.48 & 92.53 & 4.86
          & 5.11 & 0.06 & 0.47 & -0.65 & 8.10 \\
        \midrule
        0.167 & \textbf{67.21} & 71.26 & \textbf{98.20} & 90.10 & 44.41
              & \textbf{5.08} & \textbf{-0.25} & \textbf{-0.07} & -0.45 & \textbf{7.32} \\
        0.33 & 66.34 & \textbf{71.39} & 97.04 & \textbf{90.19} & 45.72
             & 4.76 & -0.32 & -0.14 & -0.50 & 6.94 \\
        0.5 & 66.03 & 69.21 & 96.75 & 88.92 & \textbf{48.68}
            & 3.97 & -0.43 & -0.25 & \textbf{-0.39} & 6.68 \\
        \bottomrule
    \end{tabular*}
    \vspace{-4pt}
\end{table*}

\begin{table*}[!ht]
    \centering
    \caption{GraftOOD ratio sweep with $n_{\mathrm{ood}}=2$ ($G=8$).
    The $\rho=0$ row is the shared FlowAWR baseline.
    Bold marks column bests among positive ratios only.}
    \label{tab:graftood_n2}
    \setlength{\tabcolsep}{3pt}
    \begin{tabular*}{\textwidth}{@{\extracolsep{\fill}}l*{4}{c}|*{6}{c}@{}}
        \toprule
        $\rho$ & AQ & IQ & SC & OC & DD & VR & VQ & MQ & TA & HPSv3 \\
        \midrule
        0 & 69.18 & 71.01 & 98.48 & 92.53 & 4.86
          & 5.11 & 0.06 & 0.47 & -0.65 & 8.10 \\
        \midrule
        0.167 & 66.59 & \textbf{71.29} & 98.18 & \textbf{90.59} & 31.41 & \textbf{5.34} & \textbf{-0.03} & \textbf{0.15} & -0.53 & \textbf{7.6} \\
        0.33 & \textbf{67.57} & 71.08 & \textbf{98.25} & 88.26 & 40.79 & 4.89 & -0.17 & 0.09 & -0.5 & 7.06 \\
        0.5 & 66.25 & 69.43 & 97.17 & 89.71 & \textbf{45.72} & 3.9 & -0.47 & -0.28 & \textbf{-0.47} & 6.54 \\
        \bottomrule
    \end{tabular*}
    \vspace{-4pt}
\end{table*}

\begin{table*}[!ht]
    \centering
    \caption{GraftOOD ratio sweep with $n_{\mathrm{ood}}=1$ ($G=8$).
    The $\rho=0$ row is the shared FlowAWR baseline.
    Bold marks column bests among positive ratios only.}
    \label{tab:graftood_n1}
    \setlength{\tabcolsep}{3pt}
    \begin{tabular*}{\textwidth}{@{\extracolsep{\fill}}l*{4}{c}|*{6}{c}@{}}
        \toprule
        $\rho$ & AQ & IQ & SC & OC & DD & VR & VQ & MQ & TA & HPSv3 \\
        \midrule
        0 & 69.18 & 71.01 & 98.48 & 92.53 & 4.86
          & 5.11 & 0.06 & 0.47 & -0.65 & 8.10 \\
        \midrule
        0.167 & \textbf{68.21} & \textbf{72.10} & 98.17 & 88.38 & 30.58 & \textbf{5.52} & -0.32 & -0.13 & \textbf{-0.39} & \textbf{7.74} \\
        0.33 & 67.43 & 71.23 & \textbf{98.27} & 90.66 & 33.93 & 5.09 & \textbf{-0.13} & \textbf{0.05} & -0.57 & 7.36 \\
        0.5 & 66.81 & 69.31 & 98.01 & \textbf{91.48} & \textbf{46.76} & 4.69 & -0.38 & -0.17 & -0.43 & 7.13 \\
        \bottomrule
    \end{tabular*}
    \vspace{-4pt}
\end{table*}

\FloatBarrier
\endgroup
\section{Limitations and Future Work}
\label{app:limitations}

\paragraph{Evaluation scope and exemplar sources.}
Our experiments are limited to text-to-video generation with Wan2.1-1.3B,
and all exemplars in GraftSet-Video are generated videos. The
$n_{\mathrm{ood}}=1$ ablation evaluates a single-exemplar group configuration,
but does not establish the effectiveness of real-video supervision. The
current results therefore leave generalization to other tasks, model scales,
and exemplar sources untested.

\paragraph{Real-video supervision and physical consistency.}
Real-world recordings could provide complementary examples of object
interactions and temporal behavior that generated exemplars may reproduce
imperfectly. Whether such supervision improves physical consistency remains
an empirical question, requiring suitable video--text pairing, data
selection, and evaluation. GraftOOD offers a regression interface for
investigating this direction because exemplar supervision requires terminal
videos rather than their generating denoising trajectories; this does not
by itself guarantee successful transfer to real-video data.

\paragraph{Extension to audio--video generation.}
A further direction is text-to-audio-video generation (T2AV), where
supervision must account for both visual dynamics and audio--visual
correspondence. Such an extension would require appropriate joint
representations and reward models sensitive to temporal synchronization
and cross-modal consistency. More broadly, advances in rewards and
evaluation remain necessary to determine whether exemplar-guided
optimization improves physical plausibility rather than merely increasing
proxy scores.

\clearpage
\section{More Qualitative Evaluation}
\label{app:more_qualitative}

\subsection{Additional Comparisons}
\label{app:qualitative_comparisons}
\begin{figure}[!t]
    \centering
    \includegraphics[width=\linewidth]{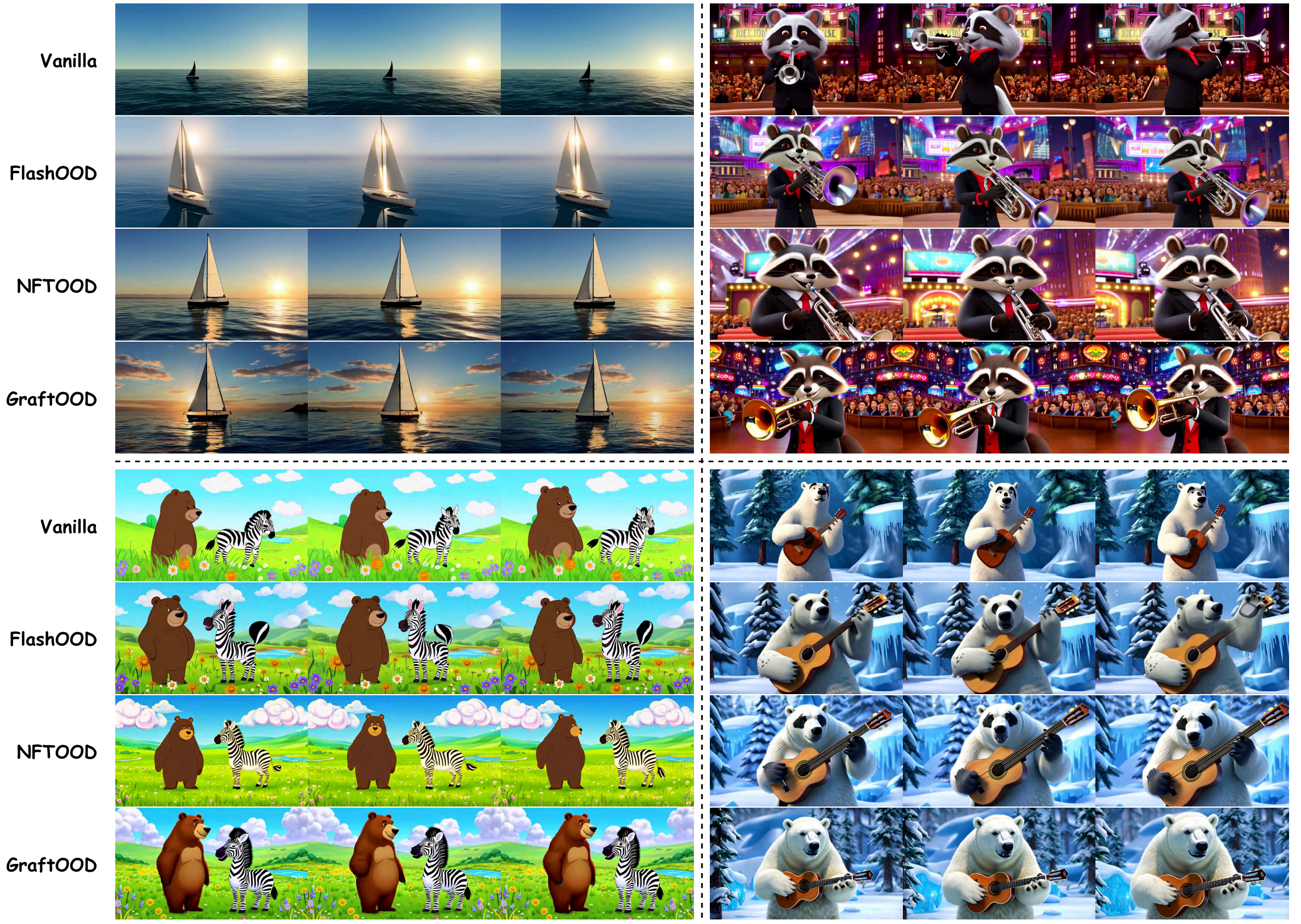}
    \vspace{-4pt}
    \caption{\textbf{Additional qualitative comparisons.}
    Each block shows Vanilla, FlashOOD, NFTOOD, and GraftOOD from top to
    bottom, with three frames in temporal order. Full prompts appear in
    Appendix~\ref{app:qualitative_prompts}.}
    \label{fig:appendix_qualitative}
    \vspace{-0.08in}
\end{figure}

We extend the qualitative comparisons in
Figures~\ref{fig:graftood_rl_compare} and~\ref{fig:graftood_compare}
with additional examples in Figure~\ref{fig:appendix_qualitative}.
We compare the vanilla generator with FlashOOD, NFTOOD, and GraftOOD.
Each example presents
three representative frames in temporal order, allowing a side-by-side
inspection of subject appearance, scene composition, and visible changes
across the sampled frames.

The additional examples cover both natural scenery and stylized animal
characters. In the sunset scene, the GraftOOD example combines a prominent
sailboat with textured clouds, warm illumination, and reflections on the
water. In the trumpet scene, it depicts a richly colored stage and audience
behind the performing raccoon. The bear-and-zebra example illustrates
differences in character rendering and meadow detail, while the polar-bear
example highlights fur, guitar, and snow-covered forest textures. Across
these selected examples, GraftOOD produces visually detailed scenes with
recognizable subjects and scene layouts across the displayed frames.
These qualitative observations complement the quantitative evaluation.

\subsection{Full Prompts for the Qualitative Comparisons}
\label{app:qualitative_prompts}

\paragraph{Prompts in Figure~\ref{fig:graftood_rl_compare}.}
The examples are listed in the order top to bottom.
\begin{enumerate}
\item a cat is running on the snow ground.

\item a man standing on top of the mountain,a scarf and hair in wind, move dolly camera to left.

\item A boat sailing leisurely along the Seine River with the Eiffel Tower in background by Vincent van Gogh.

\item 20 years old anime girl with long dark hair nice body sitting in the woods and the hair is waving with the wind, windy, lighting, warm light, sun, sun rays.

\end{enumerate}

\paragraph{Prompts in Figure~\ref{fig:graftood_compare}.}
The examples are listed in the order top-left, top-right, bottom-left, and bottom-right.
\begin{enumerate}
\item In late autumn inside a greenhouse, a gardener in a soft cardigan and floral apron fills the foreground in cinematic 4K. She trims a rose, then mists the leaves as the camera moves close to droplets hanging from petals. Outside the glass, rain falls and trees turn bronze, while inside warm light makes everything glow. Crisp textures, true tones, and gentle film grain create a cozy, intimate beauty.

\item A scenic photograph capturing the moment a steam train departs from the Glenfinnan Viaduct, a historic railway bridge in Scotland. The train moves gracefully over the arch-covered viaduct, its smoke billowing into the air. The landscape is lush with greenery, and towering rocky mountains frame the scene, creating a picturesque backdrop. The sky is a clear, bright blue with the sun shining down, casting a warm glow on the train and the surrounding scenery. The viaduct itself is a striking feature, with intricate ironwork and a verdant setting. The photo has a classic, nostalgic feel, emphasizing the natural beauty and historical charm of the location. A wide-angle shot from a slightly elevated angle, capturing both the train and the expansive landscape.

\item A cute raccoon playing guitar in a boat on the ocean.

\item A green dinosaur and a yellow kitten are playing basketball on a football field.

\end{enumerate}

\paragraph{Prompts in Figure~\ref{fig:appendix_qualitative}.}
The examples are listed in the order top-left, top-right, bottom-left, and bottom-right.
\begin{enumerate}
\item Crear un velero de 50 pies hiper realista navegando hacia la pusta de sol.

\item A playful raccoon dressed in a stylish black suit with a bright red bow tie, playing a shiny silver trumpet on a bustling city stage. The raccoon has expressive brown eyes and a charming smile, standing confidently in front of a large, colorful backdrop featuring vibrant lights, a busy crowd, and a neon sign advertising the concert. The raccoon's fur is sleek and glossy, highlighting its playful nature as it takes center stage, surrounded by enthusiastic audience members. The stage background includes intricate patterns and detailed architectural elements, creating a lively and dynamic atmosphere. Close-up, medium shot, and wide shot angles capturing the raccoon's performance from various perspectives. Smooth, vibrant animation style.

\item A whimsical wildlife scene captured in a playful animated style, featuring a majestic brown bear and a vibrant black and white zebra standing side-by-side. The bear stands tall with a gentle smile, its fur soft and fluffy, while the zebra looks alert and curious with its striped mane blowing in the breeze. They share a moment of quiet companionship in a lush green meadow dotted with wildflowers. The sky above is a vivid shade of azure with fluffy cotton candy clouds. The background showcases a serene landscape with rolling hills and a babbling brook nearby. The entire scene is rendered in a lively hand-drawn animation style with bright, cheerful colors. Close-up shot of the two animals facing each other, medium shot of their interaction, and a panoramic view of the meadow.

\item CG fantasy digital art, a majestic polar bear with a fluffy white coat and large black patches on its face, standing on its hind legs, holding a small acoustic guitar with a wooden body and intricate carvings. The bear's fur is meticulously detailed, with soft highlights and shadows. It is surrounded by a snow-covered forest with tall pine trees and sparkling ice formations. The bear is playing a lively melody, its eyes focused intently on the strings. The atmosphere is magical and enchanting. Low-angle view, focusing on the bear's expressive face and the intricate details of the guitar.

\end{enumerate}

\end{document}

%% file: math_commands.tex
\usepackage{amsmath,amsfonts,bm}

\def\eqref#1{equation~\ref{#1}}
\def\Eqref#1{Equation~\ref{#1}}

\def\1{\bm{1}}

\DeclareMathAlphabet{\mathsfit}{\encodingdefault}{\sfdefault}{m}{sl}
\SetMathAlphabet{\mathsfit}{bold}{\encodingdefault}{\sfdefault}{bx}{n}



%% file: iclr2027_conference.bib
@inproceedings{black2024ddpo,
 author = {Black, Kevin and Janner, Michael and Du, Yilun and Kostrikov, Ilya and Levine, Sergey},
 booktitle = {International Conference on Learning Representations},
 pages = {4965--4987},
 title = {Training Diffusion Models with Reinforcement Learning},
 volume = {2024},
 year = {2024}
}

@article{zheng2025diffusionnft,
  title={Diffusionnft: Online diffusion reinforcement with forward process},
  author={Zheng, Kaiwen and Chen, Huayu and Ye, Haotian and Wang, Haoxiang and Zhang, Qinsheng and Jiang, Kai and Su, Hang and Ermon, Stefano and Zhu, Jun and Liu, Ming-Yu},
  journal={arXiv preprint arXiv:2509.16117},
  year={2025}
}

@article{fan2023dpok,
  title={Dpok: Reinforcement learning for fine-tuning text-to-image diffusion models},
  author={Fan, Ying and Watkins, Olivia and Du, Yuqing and Liu, Hao and Ryu, Moonkyung and Boutilier, Craig and Abbeel, Pieter and Ghavamzadeh, Mohammad and Lee, Kangwook and Lee, Kimin},
  journal={Advances in Neural Information Processing Systems},
  volume={36},
  pages={79858--79885},
  year={2023}
}

@article{he2026flash,
  title={Flash-GRPO: Efficient Alignment for Video Diffusion via One-Step Policy Optimization},
  author={He, Xiaoxuan and Fu, Siming and Xue, Zeyue and Wang, Weijie and He, Ruizhe and Li, Yuming and Yin, Dacheng and Dong, Shuai and Huang, Haoyang and Wang, Hongfa and others},
  journal={arXiv preprint arXiv:2605.15980},
  year={2026}
}

@article{fu2026flowawr,
  title={FlowAWR: Online Adaptive Flow Reinforcement via Advantage-Weighted Rectification},
  author={Fu, Zheming and He, Ruizhe and Shang, Wei and Ma, Xiaoxiao and Wang, Lei and Liu, Chang and Fu, Siming},
  journal={arXiv preprint arXiv:2606.30376},
  year={2026}
}

@article{liu2026flow,
  title={Flow-grpo: Training flow matching models via online rl},
  author={Liu, Jie and Liu, Gongye and Liang, Jiajun and Li, Yangguang and Liu, Jiaheng and Wang, Xintao and Wan, Pengfei and Zhang, Di and Ouyang, Wanli},
  journal={Advances in neural information processing systems},
  volume={38},
  pages={40783--40818},
  year={2026}
}

@inproceedings{xu2026visionreward,
  title={Visionreward: Fine-grained multi-dimensional human preference learning for image and video generation},
  author={Xu, Jiazheng and Huang, Yu and Cheng, Jiale and Yang, Yuanming and Xu, Jiajun and Wang, Yuan and Duan, Wenbo and Yang, Shen and Jin, Qunlin and Li, Shurun and others},
  booktitle={Proceedings of the AAAI Conference on Artificial Intelligence},
  volume={40},
  number={13},
  pages={11269--11277},
  year={2026}
}

@inproceedings{teed2020raft,
  title={{RAFT}: Recurrent All-Pairs Field Transforms for Optical Flow},
  author={Teed, Zachary and Deng, Jia},
  booktitle={European Conference on Computer Vision},
  pages={402--419},
  year={2020}
}

@inproceedings{huang2024vbench,
  title={Vbench: Comprehensive benchmark suite for video generative models},
  author={Huang, Ziqi and He, Yinan and Yu, Jiashuo and Zhang, Fan and Si, Chenyang and Jiang, Yuming and Zhang, Yuanhan and Wu, Tianxing and Jin, Qingyang and Chanpaisit, Nattapol and others},
  booktitle={Proceedings of the IEEE/CVF Conference on Computer Vision and Pattern Recognition},
  pages={21807--21818},
  year={2024}
}

@inproceedings{lipman2022flow,
  title={Flow matching for generative modeling},
  author={Lipman, Yaron and Chen, Ricky TQ and Ben-Hamu, Heli and Nickel, Maximilian and Le, Matthew},
  booktitle={The eleventh international conference on learning representations},
  year={2022}
}

@inproceedings{liu2023flow,
  title={Flow straight and fast: Learning to generate and transfer data with rectified flow},
  author={Liu, Xingchao and Gong, Chengyue and Liu, Qiang},
  booktitle={International conference on learning representations (ICLR)},
  year={2023}
}

@article{shao2024deepseekmath,
  title={Deepseekmath: Pushing the limits of mathematical reasoning in open language models},
  author={Shao, Zhihong and Wang, Peiyi and Zhu, Qihao and Xu, Runxin and Song, Junxiao and Bi, Xiao and Zhang, Haowei and Zhang, Mingchuan and Li, YK and Wu, Yang and others},
  journal={arXiv preprint arXiv:2402.03300},
  year={2024}
}

@inproceedings{liu2026improving,
  title={Improving video generation with human feedback},
  author={Liu, Jie and Liu, Gongye and Liang, Jiajun and Yuan, Ziyang and Liu, Xiaokun and Zheng, Mingwu and Wu, Xiele and Wang, Qiulin and Xia, Menghan and Wang, Xintao and others},
  journal={Advances in Neural Information Processing Systems},
  volume={38},
  pages={82155--82192},
  year={2026}
}

@inproceedings{he2024videoscore,
  title={Videoscore: Building automatic metrics to simulate fine-grained human feedback for video generation},
  author={He, Xuan and Jiang, Dongfu and Zhang, Ge and Ku, Max and Soni, Achint and Siu, Sherman and Chen, Haonan and Chandra, Abhranil and Jiang, Ziyan and Arulraj, Aaran and others},
  booktitle={Proceedings of the 2024 Conference on Empirical Methods in Natural Language Processing},
  pages={2105--2123},
  year={2024}
}

@article{he2025videoscore2,
  title={Videoscore2: Think before you score in generative video evaluation},
  author={He, Xuan and Jiang, Dongfu and Nie, Ping and Liu, Minghao and Jiang, Zhengxuan and Su, Mingyi and Ma, Wentao and Lin, Junru and Ye, Chun and Lu, Yi and others},
  journal={arXiv preprint arXiv:2509.22799},
  year={2025}
}

@inproceedings{tong2025mjvideo,
  title={Mj-video: Fine-grained benchmarking and rewarding video preferences in video generation},
  author={Tong, Haibo and Wang, Zhaoyang and Chen, Zhaorun and Ji, Haonian and Qiu, Shi and Han, Siwei and Geng, Kexin and Xue, Zhongkai and Zhou, Yiyang and Xia, Peng and others},
  journal={arXiv preprint arXiv:2502.01719},
  year={2025}
}

@article{wan2025,
  title={Wan: Open and advanced large-scale video generative models},
  author={Wan, Team and Wang, Ang and Ai, Baole and Wen, Bin and Mao, Chaojie and Xie, Chen-Wei and Chen, Di and Yu, Feiwu and Zhao, Haiming and Yang, Jianxiao and others},
  journal={arXiv preprint arXiv:2503.20314},
  year={2025}
}

@article{liu2024videodpo,
  title={Videodpo: Omni-preference alignment for video diffusion generation},
  author={Liu, Runtao and Wu, Haoyu and Ziqiang, Zheng and Wei, Chen and He, Yingqing and Pi, Renjie and Chen, Qifeng},
  journal={arXiv preprint arXiv:2412.14167},
  year={2024}
}

@article{ouyang2022training,
  title={Training language models to follow instructions with human feedback},
  author={Ouyang, Long and Wu, Jeffrey and Jiang, Xu and Almeida, Diogo and Wainwright, Carroll and Mishkin, Pamela and Zhang, Chong and Agarwal, Sandhini and Slama, Katarina and Ray, Alex and others},
  journal={Advances in Neural Information Processing Systems},
  volume={35},
  pages={27730--27744},
  year={2022}
}

@article{li2025mixgrpo,
  title={Mixgrpo: Unlocking flow-based grpo efficiency with mixed ode-sde},
  author={Li, Junzhe and Cui, Yutao and Huang, Tao and Ma, Yinping and Fan, Chun and Yang, Miles and Zhong, Zhao},
  journal={arXiv preprint arXiv:2507.21802},
  year={2025}
}

@article{christiano2017deep,
  title={Deep reinforcement learning from human preferences},
  author={Christiano, Paul F and Leike, Jan and Brown, Tom and Martic, Miljan and Legg, Shane and Amodei, Dario},
  journal={Advances in neural information processing systems},
  volume={30},
  year={2017}
}

@article{xue2025dancegrpo,
  title={Dancegrpo: Unleashing grpo on visual generation},
  author={Xue, Zeyue and Wu, Jie and Gao, Yu and Kong, Fangyuan and Zhu, Lingting and Chen, Mengzhao and Liu, Zhiheng and Liu, Wei and Guo, Qiushan and Huang, Weilin and others},
  journal={arXiv preprint arXiv:2505.07818},
  year={2025}
}

@inproceedings{ma2025hpsv3,
  title={Hpsv3: Towards wide-spectrum human preference score},
  author={Ma, Yuhang and Wu, Xiaoshi and Sun, Keqiang and Li, Hongsheng},
  booktitle={Proceedings of the IEEE/CVF International Conference on Computer Vision},
  pages={15086--15095},
  year={2025}
}

@article{team2025longcat,
  title={Longcat-video technical report},
  author={Team, Meituan LongCat and Cai, Xunliang and Huang, Qilong and Kang, Zhuoliang and Li, Hongyu and Liang, Shijun and Ma, Liya and Ren, Siyu and Wei, Xiaoming and Xie, Rixu and others},
  journal={arXiv preprint arXiv:2510.22200},
  year={2025}
}

@article{team2025kling,
  title={Kling-omni technical report},
  author={Team, Kling and Chen, Jialu and Ci, Yuanzheng and Du, Xiangyu and Feng, Zipeng and Gai, Kun and Guo, Sainan and Han, Feng and He, Jingbin and He, Kang and others},
  journal={arXiv preprint arXiv:2512.16776},
  year={2025}
}

@article{seedance2025seedance,
  title={Seedance 1.5 pro: A native audio-visual joint generation foundation model},
  author={Seedance, Team and Chen, Heyi and Chen, Siyan and Chen, Xin and Chen, Yanfei and Chen, Ying and Chen, Zhuo and Cheng, Feng and Cheng, Tianheng and Cheng, Xinqi and others},
  journal={arXiv preprint arXiv:2512.13507},
  year={2025}
}

@article{seedance2026seedance,
  title={Seedance 2.0: Advancing video generation for world complexity},
  author={Seedance, Team and Chen, De and Chen, Liyang and Chen, Xin and Chen, Ying and Chen, Zhuo and Chen, Zhuowei and Cheng, Feng and Cheng, Tianheng and Cheng, Yufeng and others},
  journal={arXiv preprint arXiv:2604.14148},
  year={2026}
}

@inproceedings{ball2023efficient,
  title={Efficient Online Reinforcement Learning with Offline Data},
  author={Ball, Philip J. and Smith, Laura and Kostrikov, Ilya and Levine, Sergey},
  booktitle={Proceedings of the 40th International Conference on Machine Learning},
  pages={1577--1594},
  publisher={PMLR},
  year={2023},
}

@article{yan2026learning,
  title={Learning to reason under off-policy guidance},
  author={Yan, Jianhao and Li, Yafu and Hu, Zican and Wang, Zhi and Cui, Ganqu and Qu, Xiaoye and Cheng, Yu and Zhang, Yue},
  journal={Advances in Neural Information Processing Systems},
  volume={38},
  pages={117157--117186},
  year={2026}
}
